\pdfoutput=1
\documentclass[letterpaper]{article} 
\usepackage{aaai2026}  
\usepackage{times}  
\usepackage{helvet}  
\usepackage{courier}  
\usepackage[hyphens]{url}  
\usepackage{graphicx} 
\usepackage{natbib}  
\usepackage{caption} 
\usepackage{algorithm}
\usepackage{algorithmic}
\usepackage{makecell}
\usepackage{enumitem}
\usepackage{subcaption}
\usepackage{array}
\usepackage{booktabs}
\usepackage{multirow}
\usepackage{amsmath}

\usepackage{newfloat}
\usepackage{listings}
\DeclareCaptionStyle{ruled}{labelfont=normalfont,labelsep=colon,strut=off} 
\floatstyle{ruled}
\newfloat{listing}{tb}{lst}{}
\floatname{listing}{Listing}
\title{CCR-Bench: A Comprehensive Benchmark for Evaluating LLMs on Complex Constraints, Control Flows, and Real-World Cases}
\author{
    Xiaona Xue\textsuperscript{\rm },
    Yiqiao Huang\textsuperscript{\rm },
    Jiacheng Li\textsuperscript{\rm },
    Yuanhang Zheng\textsuperscript{\rm },
    Huiqi Miao\textsuperscript{\rm },
    Yunfei Ma\textsuperscript{\rm },
    Rui Liu\textsuperscript{\rm },
    Xinbao Sun\textsuperscript{\rm },
    Minglu Liu\textsuperscript{\rm },
    Fanyu Meng\textsuperscript{\rm },
    Chao Deng\textsuperscript{\rm },
    Junlan Feng\textsuperscript{\rm }
}
\affiliations{
    
    \textsuperscript{\rm } Jiutian Artificial Intelligence Research Institute, China Mobile, Beijing, China\\

    {\{xuexiaona,huangyiqiao,lijiacheng,mayunfei\}}@cmjt.chinamobile.com\\

}

\usepackage{bibentry}

\begin{document}

\maketitle

\begin{abstract}

Enhancing the ability of large language models (LLMs) to follow complex instructions is critical for their deployment in real-world applications. However, existing evaluation methods often oversimplify instruction complexity as a mere additive combination of atomic constraints, failing to adequately capture the high-dimensional complexity arising from the intricate interplay of content and format, logical workflow control, and real-world applications. This leads to a significant gap between current evaluation practices and practical demands. To bridge this gap, we introduce CCR-Bench, a novel benchmark designed to assess LLMs' adherence to complex instructions. CCR-Bench is characterized by: (1) deep entanglement of content and formatting requirements in task specifications; (2) instructions that involve intricate task decomposition, conditional reasoning, and procedural planning; and (3) evaluation samples derived entirely from real-world industrial scenarios. Extensive experiments on CCR-Bench demonstrate that even state-of-the-art models exhibit substantial performance deficiencies, clearly quantifying the gap between current LLM capabilities and the demands of real-world instruction understanding. We believe that CCR-Bench offers a more rigorous and realistic evaluation framework, advancing the development of LLMs toward the next generation of models capable of understanding and executing complex tasks in industrial applications.

\end{abstract}


\section{Introduction}
Large Language Models (LLMs) have emerged as foundational technologies driving advancements in artificial intelligence. Their remarkable zero-shot and few-shot generalization, derived from pretraining on extensive text corpora, is transforming natural language processing and its downstream applications~\cite{zhang2024aquila2}. Under this new paradigm, instruction following is no longer merely one of many capabilities, but rather constitutes the core mechanism for models to collaborate with humans and accomplish complex tasks. This ability directly influences how well model behavior aligns with user intent, making its reliability critical for safe and effective deployment in real-world scenarios. In high-stakes domains such as healthcare, finance, and autonomous systems, misinterpretation or deviation from instructions can lead to serious consequences. As a result, improving the precision and reliability of LLMs' instruction-following capabilities is a primary focus for both academia and industry.


Evaluating the instruction-following ability of LLMs is a crucial research focus~\cite{sun2023evaluating}. To facilitate objective and quantifiable assessment, researchers concentrated on ``verifiable instructions'', such as IFEval ~\cite{IFEval} offers reliable programmatic evaluation baselines. As research advanced, it became evident that single, discrete constraints do not capture the complexity of real-world instructions, which often involve multiple, interwoven constraints. Subsequent studies explored more intricate scenarios. For instance, FollowBench~\cite{FollowBench} probes model capability boundaries by incrementally increasing constraints. WildIFEval~\cite{lior2025wildifeval} extracts ``wild'' multi-constraint instructions from real user queries. Furthermore, works such as CFBench~\cite{CFBench} and ComplexBench~\cite{wang2025complexbench} further explore the constraint typologies, their combinatorial interactions, and hierarchical structures.


However, existing methods mainly increase instruction complexity by linearly combining atomic, verifiable constraints. They often fail to sufficiently capture the intrinsic, high-dimensional nature. While this research paradigm increases the number of constraints, it falls short of the demands of many high-value tasks in the real world, exhibiting the following specific shortcomings: (i) Existing studies typically treat content and format constraints as independent elements. However, in many advanced tasks, content and format are deeply intertwined, with format itself being an integral part of the content. (ii) Current benchmarks do not adequately assess models' capabilities in decomposing complex tasks, planning steps, making conditional judgments, controlling workflows, invoking tools, or handling nested processes; (iii) Most evaluation suites are derived from general domains or simplified scenarios, failing to fully simulate the complexity of real-world industrial scenarios that require integrating domain-specific knowledge, adhering to specific workflows or logic, and resolving content contradictions.

To address these research gaps and advance LLMs from ``following simple constraint combinations'' to ``understanding and executing truly complex tasks'', we introduce a novel complex instruction-following benchmark, named CCR-Bench. CCR-Bench comprises 174 samples, all derived from or closely simulating real industrial applications and advanced cognitive activities. It requires models to demonstrate deep integration of content and format, precise control over complex logical workflows, and profound understanding of domain-specific contexts.

%
To probe the advanced instruction-following capabilities of mainstream models, we evaluate five open-source and three closed-source models on CCR-Bench using both non-thinking and thinking reasoning modes. Experiments reveal the following: (i) Although closed-source models generally outperform open-source models, only Gemini-2.5-Pro~\cite{Gemini-2.5-Pro} surpasses the ``passing threshold'' in TSR on the logical workflow control task. In contrast, none of the models achieve an HSR score above 0.5 in the other two scenarios, indicates that current models still fall short of meeting the requirements for real-world industrial applications and underscores the challenging nature of CCR-Bench; (ii) Evaluations on CCR-Bench reveal the upper limits and shortcomings of current LLMs' instruction-following capabilities, providing directions for their future development.

\section{Related Work}

\subsection{Instruction Following}
Early research demonstrated that fine-tuning LLMs on annotated instruction datasets significantly enhances their ability to follow general language instructions. To address the complexity and diversity of user intents in real-world scenarios, research has shifted toward increasing the complexity of instruction to improve the generalization and performance of the model. For instance, ~\cite{Orca} augmented instruction complexity by generating reasoning steps for simple instructions, while ~\cite{WizardLM} explored strategies for evolving simple instructions into complex ones through ``breadth'' and ``depth''.

Moreover, recent efforts have begun to investigate instruction compositionality, i.e., how multiple constraints are combined in specific structures, such as chained, conditional, or nested configurations. Such structural complexity is a hallmark of real-world tasks, but remains underexplored in instruction-following research. Our study fills this gap by focusing on how models perform when handling authentic instructions with intricate structures and multiple constraints.

\subsection{Evaluation Methods}

Evaluating the instruction-following capabilities of LLMs is critical, and current mainstream evaluation methods can be broadly categorized into three types:

\paragraph{Human Evaluation} Human evaluators directly assess whether a model’s output aligns with the given instructions. This approach offers nuanced judgments, particularly for complex or open-ended instructions. However, it is costly and time-consuming.

\paragraph{Rule/Script-Based Evaluation} This approach leverages predefined rules or programs to automatically verify whether the model output satisfies verifiable constraints in the instruction. It offers high standardization and scalability while avoiding human or model-induced bias. For example, IFEval ~\cite{IFEval} evaluates a model's adherence to instruction constraints using pre-designed rule scripts, while RULES ~\cite{RULES} determines if a model violates any rules during a conversation through programmed functions.


\paragraph{Model-Based Evaluation} 
The method employs advanced LLMs to evaluate whether a model’s output satisfies the requirements. For instance, benchmarks like CFBench~\cite{CFBench}, InFoBench~\cite{INFOBENCH}, IOPO~\cite{IOPO} and MT-Bench~\cite{MT-Bench_and_Chatbot_Arena} utilize GPT-4 series models as evaluators. Although efficient, the reliability of this method is highly dependent on the fairness, accuracy, and consistency of the judge model. 
In order to mitigate the biases introduced by purely model-based evaluations, FollowBench~\cite{FollowBench}, ComplexBench~\cite{wang2025complexbench}, and IHEval~\cite{IHEval} adopt a hybrid evaluation approach that combines model evaluations with rule-based assessments.


\subsection{Evaluation Limitations}

Despite notable progress in instruction-following evaluation, current benchmarks exhibit several limitations: (i) They incorporate diverse or multi-level constraints that are often simple and fragmented, lacking the structural complexity found in real-world scenarios. Moreover, they fail to cover realistic application contexts systematically. (ii) They typically assess the overall quality of responses, making it difficult to isolate instruction-following performance from other factors, such as factual correctness. 
Our work aims to develop a more refined evaluation framework that specifically measures a model’s ability to follow instructions, while also considering instruction complexity and authenticity to better reflect real user needs.



\label{subsec:limitation_benchmarks}


\section{Methodology}

As previously discussed, although existing instruction-following benchmarks have significantly contributed to advancing LLMs' capabilities, particularly in language understanding, content generation, and basic instruction compliance, they still exhibit limitations in simulating complex, real-world demands. These limitations are primarily reflected in their insufficient consideration of instruction complexity, dynamism, and domain specificity.

To address these shortcomings, we construct a novel evaluation dataset named \textbf{CCR-Bench}, designed to assess LLMs' ability to follow complex instructions through a progressive and multi-dimensional framework. It comprises the following three core components:

\begin{enumerate}[leftmargin=*, label=\textbullet]
  \item \textbf{Complex Content-Format Constraints} We introduce a set of tightly coupled ``content-format'' instructions, where the content and format are intrinsically linked. These instructions require models to generate specific content while strictly adhering to predefined format constraints.
%
  \item \textbf{Logical Workflow Control} We design tasks that demand multi-turn interaction, procedural planning, and state tracking for evaluating a model’s capacity to transition from passively following instructions to actively orchestrating and executing complex workflows.
  
  
  \item \textbf{Industrial Applications} We construct evaluation data based on real-world industrial scenarios to measure the instruction-following and problem-solving capabilities of current models in practical applications. These data encompass constraints on content, format, and logical reasoning, in addition to domain-specific demands.
\end{enumerate}

This chapter introduces a hierarchical evaluation dataset and framework that progresses from ``static regulation'' to ``dynamic workflow'' and ultimately to ``comprehensive application''. The following sections detail the construction process for all the three CCR-Bench components.


\subsection{Complex Content-Format Constraints}
\label{subsec:3.1}

\begin{figure*}[t]
	\centering
	\includegraphics[width=16.5cm]{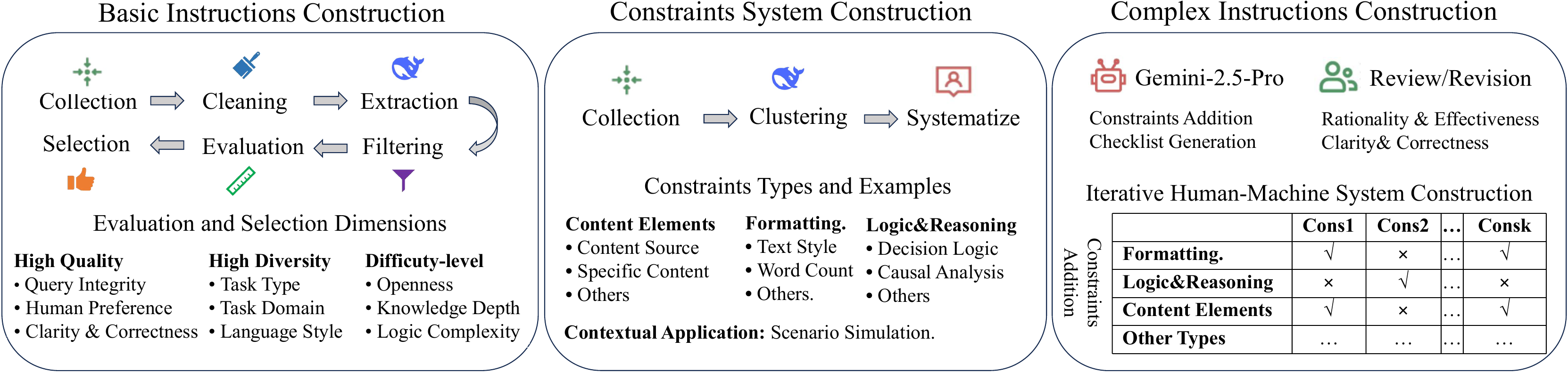}
	\caption{Framework for Complex Instructions Generation.}
	\label{fig:part1_construct}
\end{figure*}

To better reflect authentic user demands, broaden task coverage, and enable fine-grained analysis of LLMs’ instruction comprehension and basic constraint adherence abilities, we propose a framework for generating complex instructions, as illustrated in Figure~\ref{fig:part1_construct}.

\paragraph{Basic Instructions Construction}
The construction of basic instructions involves four key steps:


\textit{\textbf{1. Data Preprocessing}} 
To ensure the diversity of instruction data, we collect a large corpus of instructions from open-source datasets, including Infinity-Instruct ~\cite{Infinity_Instruct}, InstructionWild ~\cite{instructionwild}, and others ~\cite{LMSYS-Chat-1M, OpenAssistant, WildChat}. We then apply rule-based filtering to remove instructions with garbled text, HTML tags, long prompts, and those with a content similarity exceeding 0.5 computing by TF-IDF ~\cite{TF-IDF} and BGE-M3 ~\cite{BGE-M3}. 

\textit{\textbf{2. Basic Instruction Extraction}} 
We leverage advanced LLMs and prompts to extract basic instructions and constraints from the dataset. The extracted basic instructions are filtered to ensure semantic integrity and excitability by: (i) removing incomplete instructions; (ii) eliminating contradictory or unanswerable questions; (iii) excluding unethical or unsafe content.

\textit{\textbf{3. Basic Instruction Evaluation}} 
We assess the quality, diversity, and difficulty level of the basic instructions through the following methods: (i) The quality is scored by the model on a scale of 0 to 10 across content completeness, clarity, and precision, and alignment with human preferences; (ii) The diversity is evaluated by task domain, task type, and language style; (iii) Basic instructions are categorized into five difficulty levels—very low, low, medium, high, and very high, based on factors such as knowledge depth, logical complexity, and openness. Experiments indicate that these instructions span over 60 common domains and encompass 11 mainstream language styles, and the task types cover 9 first-level categories and 34 second-level subcategories. The details are displayed in Appendix A.1.1.

\textit{\textbf{4. Basic Instruction Selection}} 
After the above process, each instruction is labeled with task type, domain, language style, quality score, and difficulty level. We select instructions with quality scores above 8,  ``high'' difficulty, and distinct values across all other labels. Following random sampling and manual validation, 80 high-quality basic instructions are finally retained for downstream use.

\paragraph{Constraints System Construction}

\begin{figure}[htbp]
  \centering 
  \includegraphics[width=0.8\columnwidth]{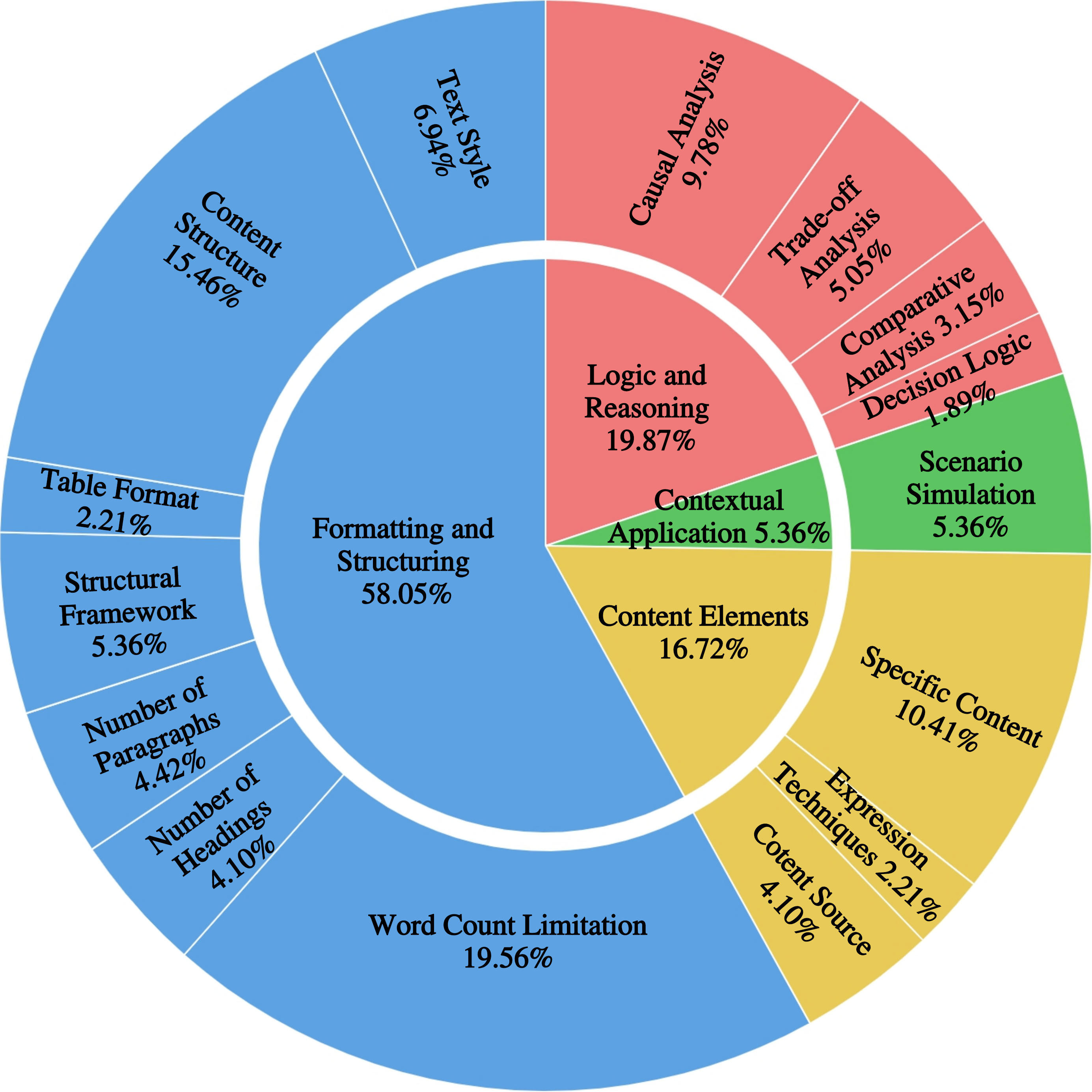} 
  \caption{Distribution of Constraints System.}
  \label{fig:part1_basic_diversity}
\end{figure}

We cluster and categorize the constraints previously extracted, identifying those that are both prevalent in practical use but prone to failure by mainstream LLMs, such as word count limits, emphasis on specific context, and tasks involving causal reasoning or trade-off analysis. Through systematic analysis, we identify and classify these constraints into four categories: Formatting and Structuring, Logic and Reasoning, Content Elements, and Contextual Application. These types, illustrated in Figure~\ref{fig:part1_basic_diversity}, represent the core objectives of our evaluation.


\paragraph{Complex Instructions Construction}


To enhance the systematicity and controllability of tasks, we utilize the Gemini-2.5-Pro model along with a carefully designed prompt to effectively integrate the above four types of constraints into the selected basic instructions. This process results in a complex instruction set that more closely reflects realistic user needs. These constraints are highly interwoven, significantly increasing the complexity and diversity of the original basic instructions.


To ensure data quality, an expert team reviews and refines the generated complex instruction set based on the following principles: content validity, clarity of intent, scenario appropriateness, data diversity and complexity. The final set consists of 64 complex instructions characterized by high quality, high diversity, and high difficulty, and each instruction contains 2 to 6 constraints.

\subsection{Logical Workflow Control}
\label{subsec:3.2}
The complexity of current mainstream instruction-following benchmarks primarily lies in imposing a set of static, one-off constraints on the model. Although these tasks require a model to integrate and satisfy multiple conditions, they are inherently stateless. They do not probe a model’s ability to decompose a task over multiple turns, to plan stepwise, to make conditional decisions, or to control the execution workflow.  In real‑world business scenarios, such as booking flights, online shopping, or conducting data analysis, the model must function as an intelligent agent capable of understanding and executing multi‑step workflows.  



To address these limitations, we draw inspiration from FlowBench~\cite{FlowBench} to construct a more realistic, application-oriented dataset of Logical  Workflow Control with automated validation scripts. This dataset is designed to capture complex scenarios, including nested workflows, implicit tool invocations, and long tool-use chains.

\paragraph{Data Construction Guidelines}

We design the following three guidelines to simulate complex interaction challenges in real-world scenarios:

\textit{\textbf{1. Workflow Structures}} 
To reflect the non-linear nature of real-world tasks, we design complex workflows with two key structures: (i) Conditional branching: The workflow path diverges based on conditions, demanding if-then-else reasoning. (ii) Implicit nesting: Sub-workflows are invoked within a main workflow and return to their origin point upon completion. Crucially, these nested workflows are not explicitly indicated, requiring the model to infer when to initiate a sub-process autonomously.

\textit{\textbf{2. Implicit Tool Invocation}} 
We design scenarios in which the model must infer missing steps from the context, as not all required actions are explicitly specified.

\textit{\textbf{3. Long‑Chain Task Planning}} 
To assess a model's planning and memory capabilities, we craft tasks that demand extended tool‑calling sequences to complete, including navigating a maze or colouring a complex tree structure. Under these circumstances, the model should adhere to each instruction step, as a single misstep leads to failure. This imposes stringent requirements on the model's robustness in long-term planning and instruction adherence.




\paragraph{Data Construction Process}

The construction pipeline is described as follows. 


\textit{\textbf{1. Environment Construction}} 
We first define a standardized and interactive execution environment comprising the following components: (i) Design a set of workflows representing diverse business scenarios. Each workflow consists of well-defined states and transition rules, formally encoding the correct task execution path; (ii) Establish a toolkit, with each tool accompanied by a specified API interface detailing its functionality, input parameters, expected output format, and the executable code. 

\textit{\textbf{2. Real-World Scenario}} 
To ensure the dataset aligns with real-world applications, we adopt an expert-driven approach to construct high-quality data across six realistic and common scenarios, such as flight booking, telecommunications services, etc. The construction process is as follows: (i) Create diverse user profiles with a clear user goal, supplementary information required for tool invocation during the dialogue, and trigger phrases that may initiate specific workflows. (ii) For each profile, experts manually author the ideal tool-invocation sequence required to accomplish the user’s task based on the predefined workflow, and then develop verification scripts that can automatically validate whether the model’s final state aligns with the intended outcome.


  
    

\textit{\textbf{3. Abstract Scenario}} 
To further investigate the model’s task-planning abilities over extended tool-invocation chains, we construct data for three abstract scenarios through the following process: based on predefined workflows and tool-sets, we use code to automatically produce test cases requiring extremely long sequences of tool invocations to resolve, and develop deterministic programs to generate the ground-truth corresponding to each instance.


Through the process described above, we construct a dataset containing 70 instances (see Table~\ref{tab:Basic Information of the Logical Control Process Dataset}). This strategy ensures both authenticity and complexity, providing comprehensive coverage to robustly evaluate the logical workflow control capabilities of LLMs.

\begin{table}[htbp]
    \centering
    \small 
    \begin{tabular}{llc}
        \toprule[0.5mm]
        \textbf{\makecell{}} & \textbf{Scenario Name} & \textbf{\makecell{Number}} \\

        \midrule[0.3mm]
        \multirow{6}{*}{\textbf{\makecell[l]{Real-World\\Scenario}}} & Data Workflow & 5 \\
        & Flight Ticket Booking & 10 \\
        & Real Estate & 10 \\
        & Customer Service & 10 \\
        & Online Game & 10 \\
        & Printer Assistant & 10 \\
        \midrule
        \multirow{3}{*}{\textbf{\makecell[l]{Abstract\\Scenario}}} & Maze & 5 \\
        & Tree Painter & 5 \\
        & World Cup Simulator & 5 \\
        \bottomrule[0.5mm]
    \end{tabular}
    \caption{Properties of the Logical Workflow Control Data.}
    \label{tab:Basic Information of the Logical Control Process Dataset}
\end{table}

\subsection{Industrial Applications}
\label{subsec:3.3}
To evaluate a model’s practicality and robustness under realistic industrial conditions, we build a novel set of integrated tasks that impose constraints on both content and format, require rigorous logical reasoning, and are tightly coupled with specific industry backgrounds. The goal is to replicate the core challenges present in real industrial applications faithfully. To ensure the authenticity, complexity, and reliability of our benchmark, we design and implement a systematic four-stage data construction pipeline as follows.


\paragraph{Data Collection}

In strict compliance with data privacy and security protocols, we collect a large volume of unfiltered, frontline user logs from the real medical scenario.

\paragraph{Data Refinement}
 To construct a high‑quality dataset, we perform meticulous preprocessing in two stages: (i) Invalid, garbled, or truncated entries are discarded initially. Subsequently, a hybrid pipeline of automated scripts and manual validation rigorously anonymizes the data. These procedures preserve data fidelity while safeguarding user privacy and ensuring legal compliance. (ii) We establish a set of filtering criteria that favour entries exhibiting high complexity, and diversity. Specifically, we prioritize instructions that impose multiple content or format constraints, span multiple task steps, embed implicit conditions or distracting information, or require non‑trivial logical reasoning.
 

\paragraph{Evaluation Dimension}
A two‑step ``LLM-assisted + human verification'' method is adopted to systematically define the evaluation dimensions: (i) We employ advanced LLMs to perform a comprehensive extraction of constraints from each instruction; (ii) Experts manually review the extracted constraints. They correct erroneous extractions, supplement missing complex constraints, and eliminate unreasonable or trivial items, abstract and synthesize the validated constraints into a final set of evaluation dimensions.


\paragraph{Evaluation Data Construction}
With the evaluation dimensions defined, we proceed to the final stage of constructing the evaluation dataset: (i) Experts conduct a comprehensive quality assessment of each data pair consisting of a complex instruction and its verified constraint list, and retain only those with clear instructions, reasonable constraints, and strong representativeness; (ii) We perform stratified sampling based on scenario categories and difficulty levels to construct a balanced and representative evaluation set; (iii) For each selected instance, domain experts manually craft an ideal reference response.



Through these four tightly integrated stages, we construct a high-quality evaluation dataset grounded in real-world applications and features, with clearly defined evaluation dimensions and explicit assessment criteria. This dataset, comprising 40 samples and 15 constraint types, provides a solid foundation for objectively measuring LLMs’ capabilities in complex industrial scenarios.

\section{Experiments}
\subsection{Experimental Setup}
\paragraph{Models}
We evaluate eight mainstream models, including GPT-4.1~\cite{GPT-4.1}, Gemini-2.5-Pro~\cite{Gemini-2.5-Pro}, OpenAI-o3-mini~\cite{OpenAI-o3-mini}, DeepSeek-V3-0324~\cite{DeepSeek-V3-0324}, DeepSeek-R1-0528~\cite{DeepSeek-R1-0528}, QwQ-32B~\cite{QwQ-32B}, Qwen3-32B~\cite{Qwen3}, and Qwen2.5-72B-Instruct~\cite{Qwen2.5-72B-Instruct}. The closed-source models like GPT-4.1, Gemini-2.5-Pro, and OpenAI-o3-mini are accessed via their respective official APIs, while the open-source models are executed locally. The evaluation covers both thinking and non-thinking modes.

\paragraph{Parameter Settings}
For our experiments, the temperature parameter for all models was set to 0, while other parameters were kept at their default values.

\subsection{Evaluation Metrics}
\label{subsec:3.2}





This study employs three categories of metrics to evaluate the model's performance on scenario-specific data, defined or described as follows:

\paragraph{Constraint Satisfaction Rate}

We evaluate constraint satisfaction using the Hard Satisfaction Rate (HSR) and Soft Satisfaction Rate (SSR), as defined in~\cite{FollowBench}. HSR focuses on assessing the model's robustness and stability in fully adhering to composite task constraints, and SSR emphasizes the model's overall understanding and localized execution capabilities across various constraints, revealing strengths or weaknesses in specific constraints.







\paragraph{Task Success Rate}
The Task Success Rate (TSR) measures whether a model executes a task exactly as specified by the instructions, achieving all designated objectives. It prioritizes flawless replication of the execution process and absolute correctness of the result.

\paragraph{Task Completion Rate}
The Task Completion Rate (TCR) quantifies the proportion of tasks completed by the model in accordance with the instructions. Across all task types, TCR is consistently defined as:
\[
\text{TCR} = \frac{\text{correct number of } a}{\text{total number of } a},
\]
where $a$ refers to tool invocations in real-world tasks, output path nodes in Maze, drawn nodes in Drawing Tree, and predicted matches in World Cup Simulation, respectively.

We employ an automated evaluation framework, which integrates LLMs with predefined rules. For the Logical Workflow Control scenario, the ``rule/script'' method is utilized to calculate TSR and TCR. Conversely, a ``model + rule'' approach is adopted to determine HSR and SSR scores in other scenarios. To ensure accuracy and objectivity, we compute each metric as the average of 10 repeated assessments of the model outputs.


\subsection{Results and Analysis}
\paragraph{Complex Content-Format Constraints}
Table~\ref{tab:Evaluation Results of Each Model on Complex Content Specification Data} presents the evaluation results of eight mainstream models on the complex content-format constraints dataset. The results indicate that models operating under the thinking mode significantly outperform those in the non-thinking mode. The highest HSR and SSR scores are both achieved in the thinking mode by OpenAI-o3-mini and DeepSeek-R1-0528, respectively. This demonstrates that incorporating a thinking mode enhances the model's ability to understand and handle single-constraint instructions. In contrast, all models exhibit poor performance under compound constraints, highlighting the current limitations of existing models in following complex instructions. 

\begin{table}[h]
  \centering
  \small 
  \begin{tabular}{llcc} 
    \toprule[0.5mm]
      & \textbf{Model} & \textbf{HSR} & \textbf{SSR} \\
    \midrule[0.3mm]
    \multirow{5}{*}{\textbf{Non-thinking}} 
    & GPT-4.1 & \textbf{0.119} & 0.702 \\
    & DeepSeek-V3-0324 & 0.097 & \textbf{0.704} \\
    & Qwen3-32B & 0.089 & 0.651 \\
    & Qwen2.5-72B-Instruct & 0.091 & 0.631 \\
    \midrule 
    \multirow{5}{*}{\textbf{Thinking}} 
    & Gemini-2.5-Pro & 0.064 & 0.758 \\
    & OpenAI-o3-mini & \textbf{0.166} & 0.755 \\
    & DeepSeek-R1-0528 & 0.158 & \textbf{0.783} \\
    & QwQ-32B & 0.122 & 0.718 \\
    & Qwen3-32B & 0.094 & 0.672 \\
    \bottomrule[0.5mm]
  \end{tabular}
  \caption{Results in Complex Content-Format Constraints.}
  \label{tab:Evaluation Results of Each Model on Complex Content Specification Data}
\end{table}

To further examine the models' adherence to different types of constraints, we analyze their performance across four categories: content elements, formatting and structuring, logic and reasoning, and contextual application. As shown in Table~\ref{tab:SSR Metric Results of Each Model Across Different Types of Constraints}, DeepSeek-R1-0528 achieves the best overall performance, followed by Gemini-2.5-Pro and OpenAI-o3-mini. A comparative analysis reveals that there exists a consistent trend across these models: they perform poorest on formatting and structuring constraints, primarily due to the constraints' frequent entanglement with content requirements (e.g., word count limits), while their capabilities on the other three constraint types remain relatively robust. These results suggest that current models still lack a deep understanding and the ability to follow complex instructions.



\begin{table*}[t] 
  \centering
  \small 
  \begin{tabular}{llcccc}
    \toprule[0.5mm]
      &
    \textbf{Model} &
    \textbf{\makecell[c]{Content \\ Elements}} &
    \textbf{\makecell[c]{Formatting and \\ Structuring}} &
    \textbf{\makecell[c]{Logic and \\ Reasoning}} &
    \textbf{\makecell[c]{Contextual \\ Application}} \\
    \midrule[0.3mm]
    \multirow{5}{*}{\textbf{Non-thinking}} & GPT-4.1 & \textbf{0.913} & 0.579 & \textbf{0.846} & \textbf{0.841} \\
    & DeepSeek-V3-0324 & 0.908 & \textbf{0.595} & 0.818 & 0.835 \\
    & Qwen3-32B & 0.874 & 0.575 & 0.654 & 0.771 \\
    & Qwen2.5-72B-Instruct & 0.800 & 0.571 & 0.630 & 0.753 \\
    \midrule
    \multirow{5}{*}{\textbf{Thinking}} & Gemini-2.5-Pro & 0.966 & 0.623 & 0.944 & 0.882 \\
    & OpenAI-o3-mini & 0.913 & \textbf{0.673} & 0.827 & 0.882 \\
    & DeepSeek-R1-0528 &\textbf{ 0.989} & 0.658 & \textbf{0.938} & \textbf{0.918} \\
    & QwQ-32B & 0.968 & 0.610 & 0.794 & 0.829 \\
    & Qwen3-32B & 0.940 & 0.586 & 0.673 & 0.765 \\
    \bottomrule[0.5mm]
  \end{tabular}
  \caption{SSR Metrics Across Different Constraint Categories.}
  \label{tab:SSR Metric Results of Each Model Across Different Types of Constraints}
\end{table*}

\paragraph{Logical Workflow Control}


Table~\ref{tab:Evaluation Results of Each Model on Logical Control Process Data} reports the TSR and TCR scores on the logical workflow control dataset. Thinking models consistently outperform non-thinking ones, models under non-thinking mode are nearly incapable of solving these tasks correctly. For instance, Qwen3-32B shows TSR and TCR gains of 22.9\% and 21.9\% under thinking mode compared to its non-thinking variant. These results indicate that the thinking mechanism enhances logical reasoning, planning, complex process handling, and instruction adherence. The detailed experiments on abstract scenarios are shown in  Appendix B.2.


\begin{table}[h]
  \centering
  \small 
  \begin{tabular}{llcc} 
    \toprule[0.5mm]
      & \textbf{Model} & \textbf{TSR} & \textbf{TCR} \\
    \midrule[0.3mm]
    \multirow{5}{*}{\textbf{Non-thinking}} 
    & GPT-4.1 & \textbf{0.529} & \textbf{0.800} \\
    & DeepSeek-V3-0324 & 0.300 & 0.562 \\
    & Qwen3-32B & 0.157 & 0.438 \\
    & Qwen2.5-72B-Instruct & 0.329 & 0.631 \\
    \midrule 
    \multirow{5}{*}{\textbf{Thinking}} 
    & Gemini-2.5-Pro & \textbf{0.700} & \textbf{0.844} \\
    & OpenAI-o3-mini & 0.514 & 0.768 \\
    & DeepSeek-R1-0528 & 0.400 & 0.644 \\
    & QwQ-32B & 0.386 & 0.693 \\
    & Qwen3-32B & 0.386 & 0.657 \\
    \bottomrule[0.5mm]
  \end{tabular}
  \caption{Results in Logical Workflow Control.}
  \label{tab:Evaluation Results of Each Model on Logical Control Process Data}
\end{table}

Further analysis reveals that models exhibit higher TCR score than TSR, especially in thinking mode. This suggests models can partially follow workflow instructions but lack robustness in extended tool scenarios, resulting in task failure. Despite demonstrating the best performance, Gemini-2.5-Pro still shows significant room for improvement in handling such tasks.



To gain deeper insight into the causes of suboptimal model performance, we analyze real-world scenario results from three aspects: tool-use chain length, nested workflows, and implicit tool invocation.

\textbf{\textit{1. Impact of Tool-Use Chain Length}} Generally, longer tool-use chains are associated with increased task complexity and execution difficulty. Therefore, we divide the real-world samples by chain length(1-3, 4-6, 7+), representing short, medium, and long chains. Figure 3 shows that the model performance decreases as the length of the tool-use chain increases.

\begin{figure}[htbp]
    \centering
    \begin{subfigure}[t]{0.49\columnwidth} 
        \centering
        \includegraphics[width=\linewidth]{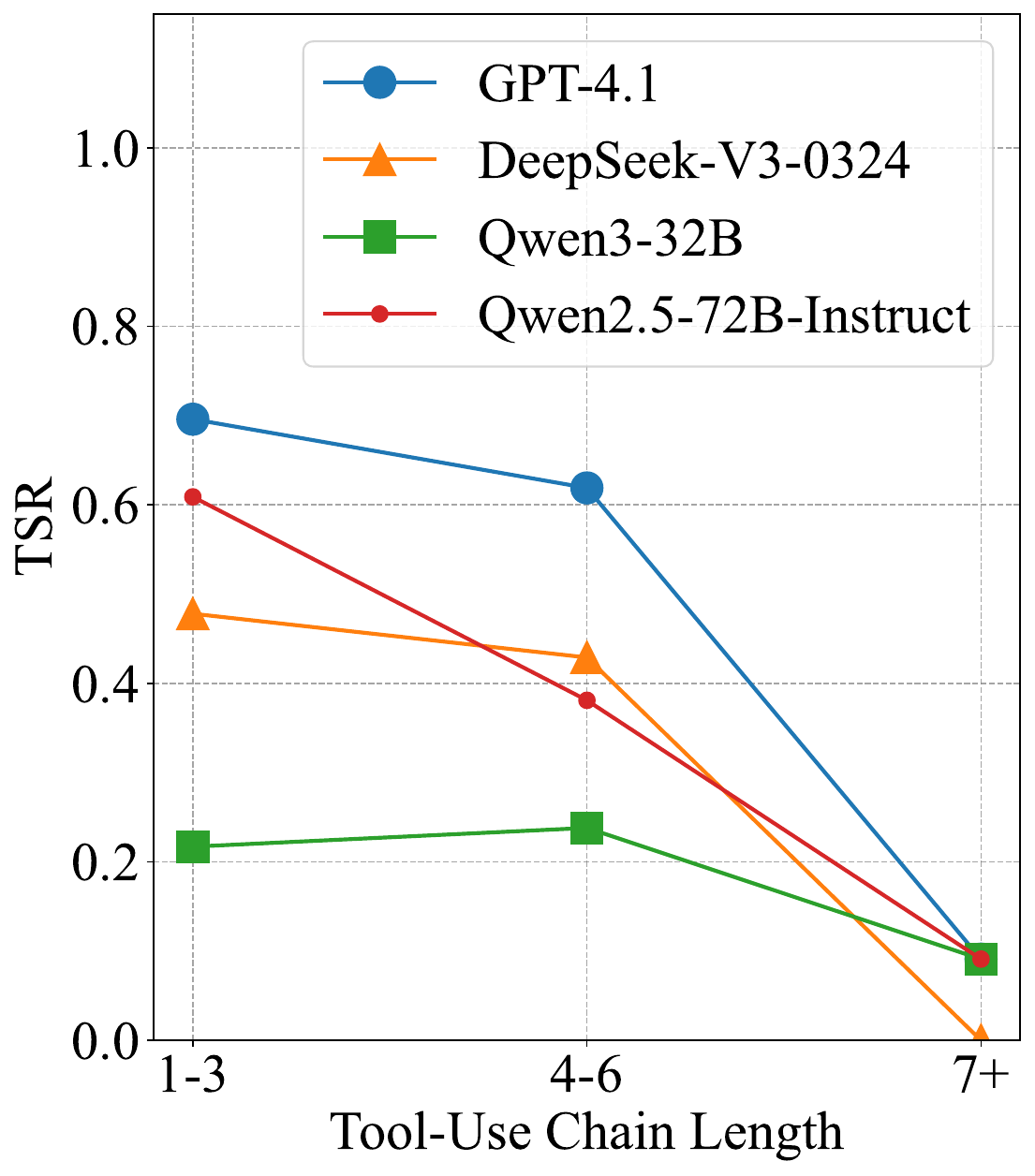}
        \caption{Non-thinking}
        \label{fig:Non_thinking_tool_chain}
    \end{subfigure}
    \hfill 
    \begin{subfigure}[t]{0.49\columnwidth} 
        \centering
        \includegraphics[width=\linewidth]{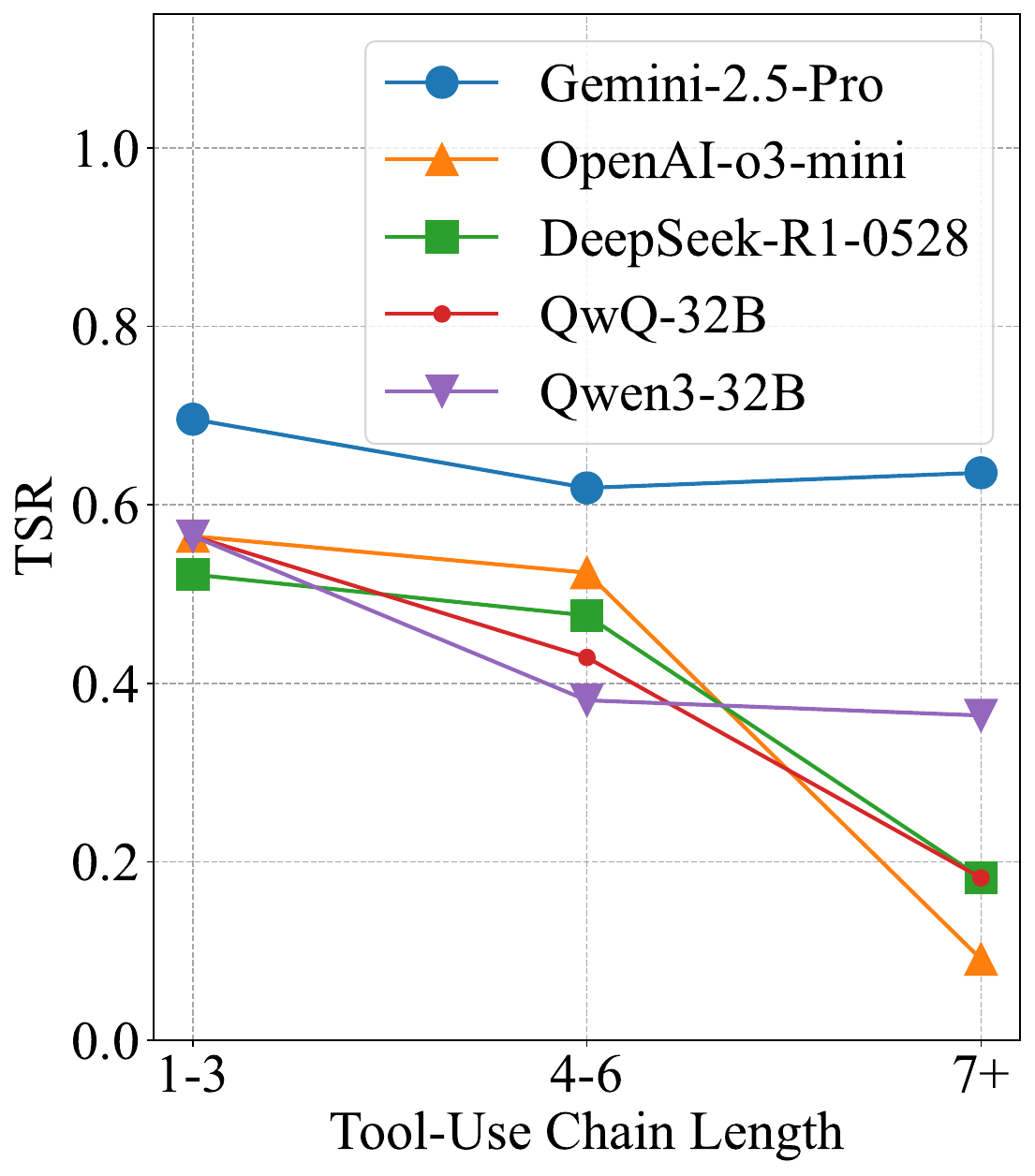}
        \caption{Thinking}
        \label{fig:Thinking-tool-chain}
    \end{subfigure}
    \caption{Impact of Tool-Use Chain Length on TSR.}
    \label{fig:part2_results_chain_length}
\end{figure}

\begin{figure}[htbp]
    \centering
    \begin{subfigure}[t]{0.49\columnwidth} 
        \centering
        \includegraphics[width=\linewidth]{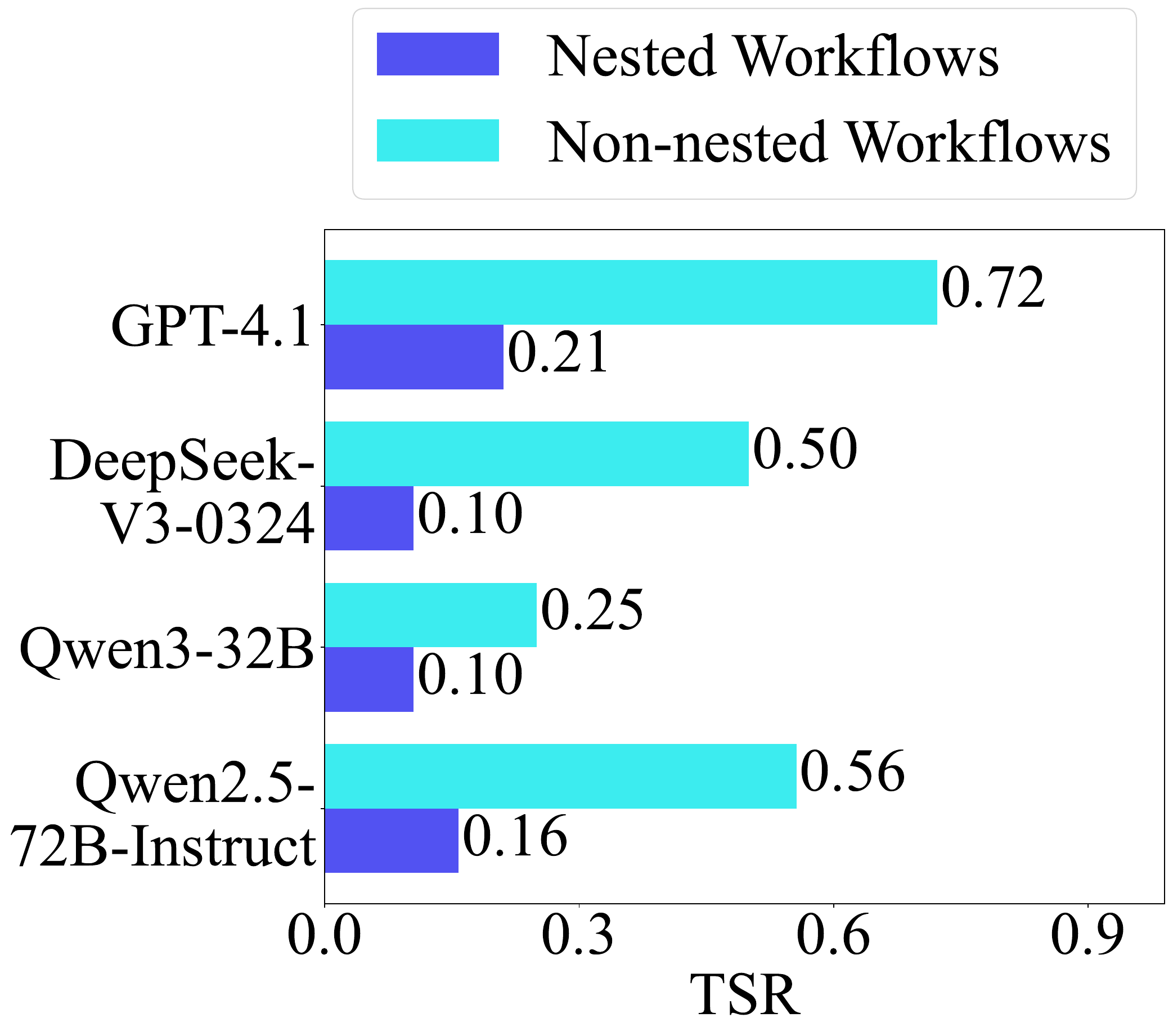}
        \caption{Non-thinking}
        \label{fig:Non_thinking-nested-processes}
    \end{subfigure}
    \hfill 
    \begin{subfigure}[t]{0.49\columnwidth} 
        \centering
        \includegraphics[width=\linewidth]{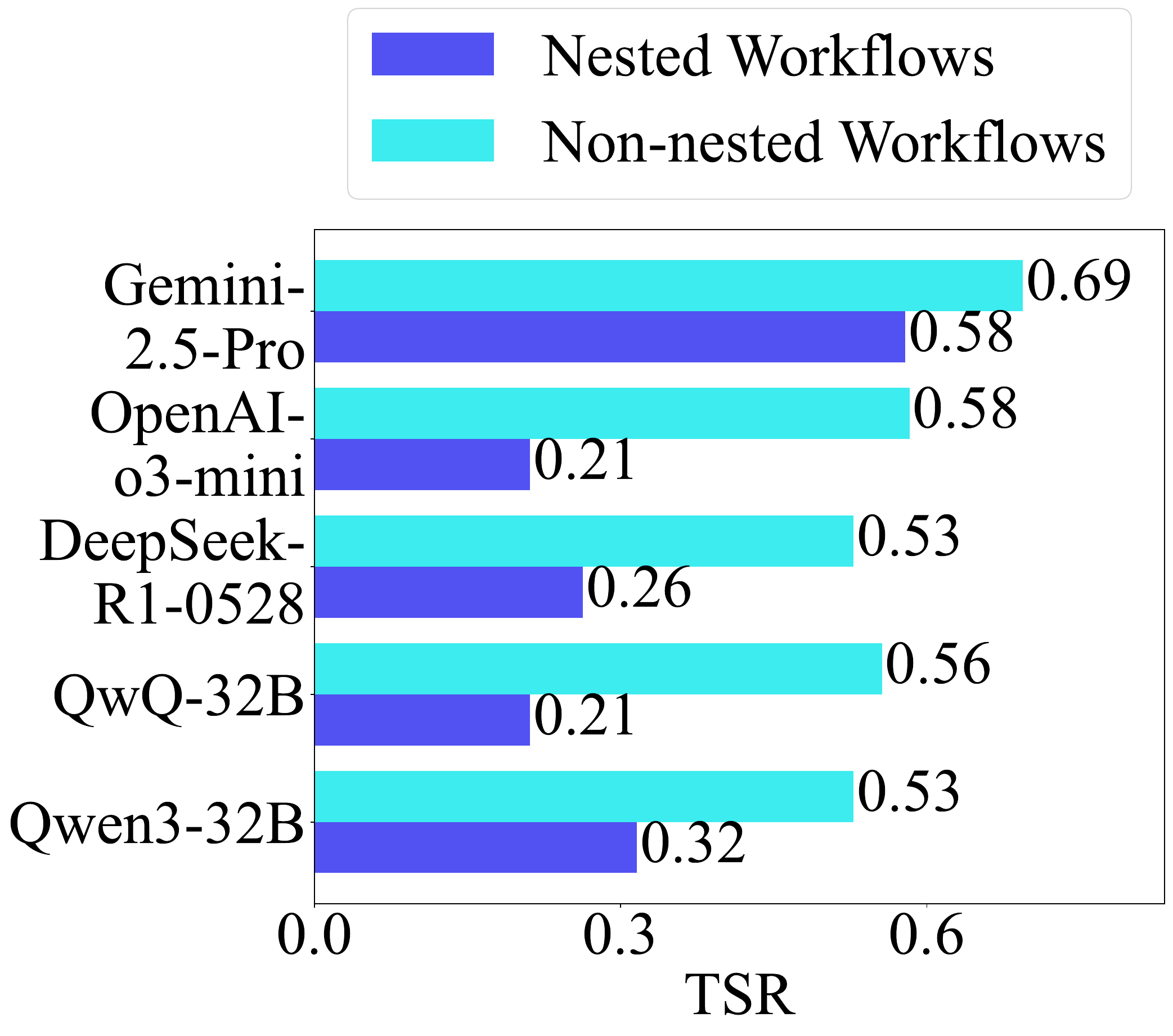}
        \caption{Thinking}
        \label{fig:Thinking-nested-processes}
    \end{subfigure}
    \caption{Impact of Nested Workflows on TSR.}
    \label{fig:part2_results_nested_workflows}
\end{figure}

\textbf{\textit{2. Impact of Nested Workflows}}
We divide the real-world scenario dataset into two categories: samples with nested workflows (19 samples) and those without (36 samples). As shown in Figure~\ref{fig:part2_results_nested_workflows}, models perform consistently worse on tasks involving nested workflows, indicating higher complexity and difficulty in execution.

\begin{figure}[htbp]
    \centering
    \begin{subfigure}[t]{0.49\columnwidth} 
        \centering
        \includegraphics[width=\linewidth]{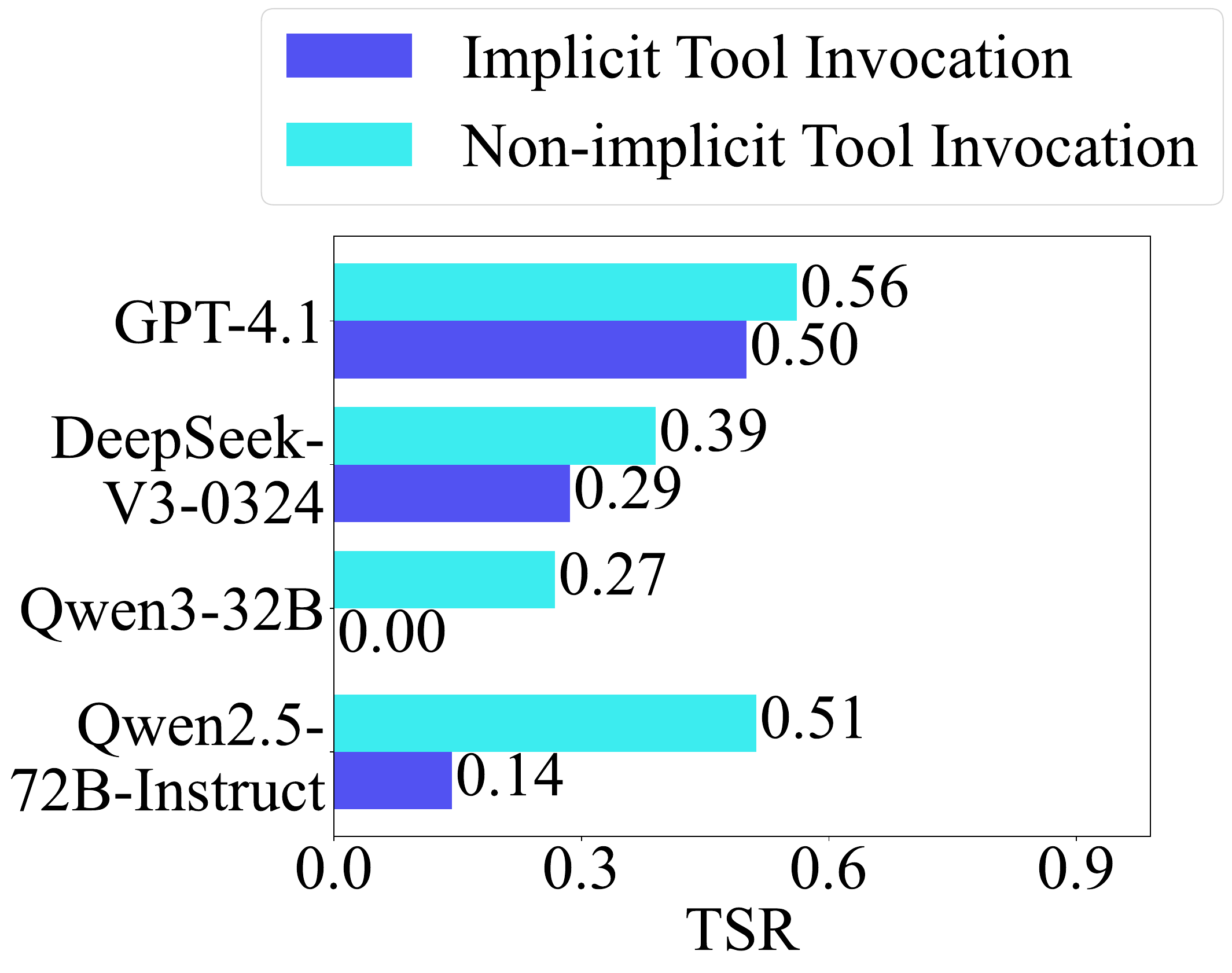}
        \caption{Non-thinking}
        \label{fig:Non_thinking-tool-calls}
    \end{subfigure}
    \hfill 
    \begin{subfigure}[t]{0.49\columnwidth} 
        \centering
        \includegraphics[width=\linewidth]{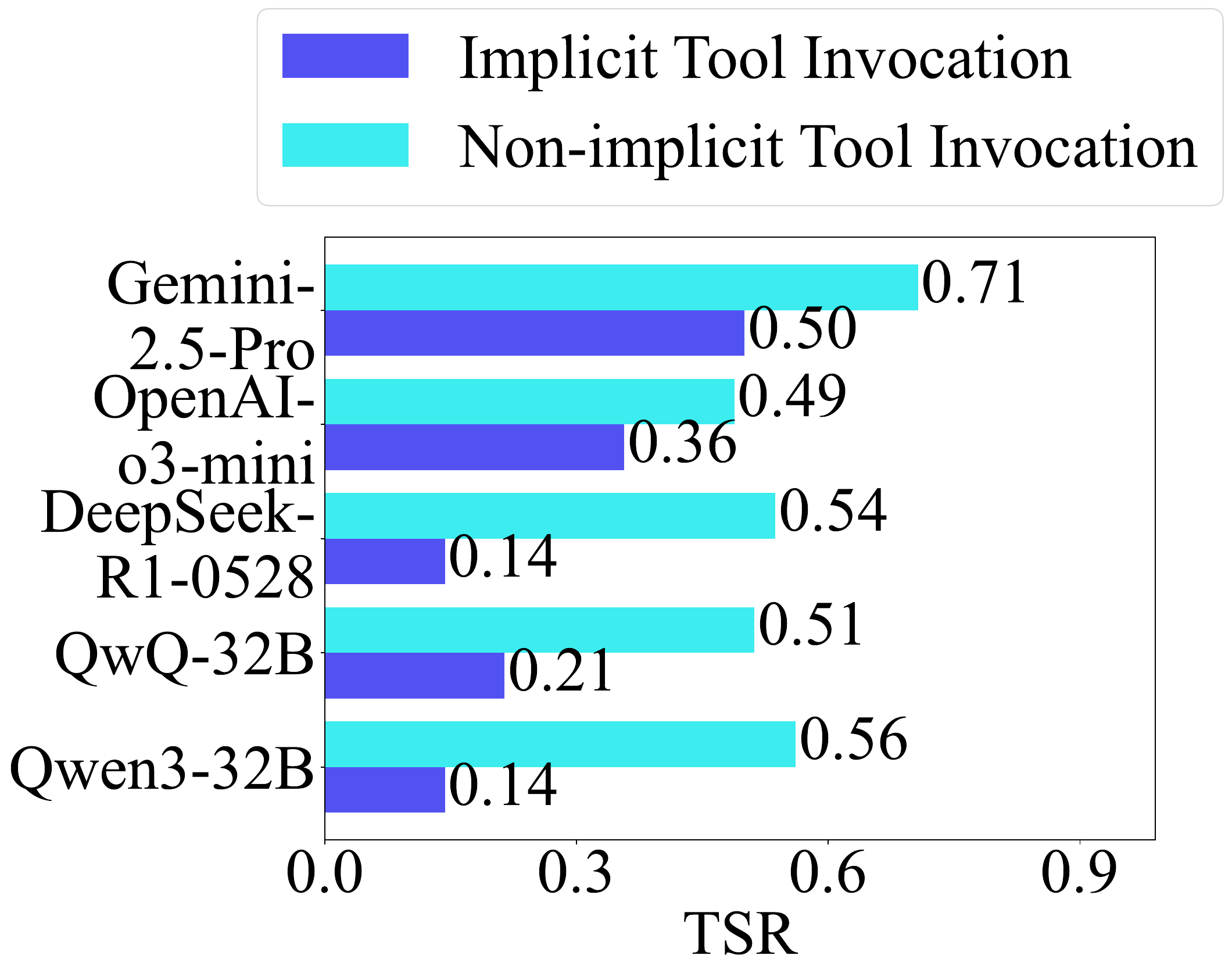}
        \caption{Thinking}
        \addtocounter{subfigure}{-1} 
        \label{fig:Thinking-tool-calls}
    \end{subfigure}
    \caption{Impact of Implicit Tool Invocation on TSR.}
    \label{fig:part2_results_implicit_tool}
\end{figure}

\textbf{\textit{3. Impact of Implicit Tool Invocation}}
We further examine model performance on tasks with implicit tool invocation (14 samples) versus those without (41 samples). Figure~\ref{fig:part2_results_implicit_tool} illustrates that tasks requiring implicit tool invocation are substantially more difficult, with lower execution success across all models.

\paragraph{Industrial Applications}

Table~\ref{tab:Evaluation Results of Different Models in Medical Scenarios} presents model performances across various tasks within industrial applications, specifically focusing on medical scenario. Observations reveal that the highest HSR and SSR scores are both achieved by the Gemini-2.5-Pro model operating in thinking mode, even though its HSR score is only 0.415.

SSR scores in this scenario generally range from 0.6 to 0.85, reflecting reasonable model competence with single constraints. However, performance drops significantly for compound constraints, highlighting the gap between current model capabilities and the demands of industrial tasks.

\begin{table}[h]
  \centering
  \small
  \begin{tabular}{llcc} 
    \toprule[0.5mm]
      & \textbf{Model} & \textbf{HSR} & \textbf{SSR} \\ 
    \midrule[0.3mm]
    \multirow{5}{*}{\textbf{Non-thinking}} 
    & GPT-4.1 & 0.135 & 0.640 \\
    & DeepSeek-V3-0324 & 0.102 & 0.647 \\
    & Qwen3-32B & \textbf{0.158} & \textbf{0.689} \\
    & Qwen2.5-72B-Instruct & 0.080 & 0.558 \\
    \midrule 
    \multirow{5}{*}{\textbf{Thinking}} 
    & Gemini-2.5-Pro & \textbf{0.415} & \textbf{0.817} \\
    & OpenAI-o3-mini & 0.242 & 0.652 \\
    & DeepSeek-R1-0528 & 0.315 & 0.721 \\
    & QwQ-32B & 0.152 & 0.610 \\
    & Qwen3-32B & 0.247 & 0.662 \\
    \bottomrule[0.5mm]
  \end{tabular}
  \caption{Results in Industrial Applications.}
  \label{tab:Evaluation Results of Different Models in Medical Scenarios}
\end{table}

 We analyze the responses of five models (See details in Appendix B.3) on specific medical example and find that most models, though partially correct, exhibit over-inference and instruction violations, leading to lower HSR scores. For example, Gemini-2.5-Pro exhibits over-inference and fails to adhere to the ``Prohibit Outputting Extra Content'' constraint. OpenAI-o3-mini failed to comply with the ``Prohibit Copying System Content'' constraint. Furthermore, models like DeepSeek-R1-0528, Qwen3-32B, and QwQ-32B violate the ``Content Filling Requirement'' constraint. These observations highlight a lack of robustness against instruction interference, significantly affecting task success. Further analysis reveals that although most models handle simple constraints reliably, they lack robustness on instructions requiring deeper comprehension, especially under ambiguity or interference. This underscores a persistent gap in aligning model behavior with nuanced user intent.

\section{Conclusion}
Instruction-following capability is a foundational aspect of a model's utility and serves as a critical link to complex real-world applications. This paper identifies a core limitation in current research on complex instruction following: the prevailing definition of ``complexity'' is overly simplistic, relying mainly on the linear accumulation of constraints, and there is a huge gap between it and real application scenarios.

To bridge this gap, we introduce CCR-Bench, a benchmark and evaluation framework designed to reflect real-world application scenarios better. CCR-Bench incorporates richer instruction structures and more intricate constraint combinations, aiming to assess the capabilities and limitations of current LLMs more accurately. Evaluation across state-of-the-art models demonstrates their limited effectiveness in high-complexity instruction scenarios, underscoring a persistent gap between model outputs and user intent.


Our work is not merely a valuable complement to existing evaluation frameworks; we envision CCR-Bench as a catalyst, driving the development of more powerful and reliable language models capable of handling the complex demands of real-world applications. Looking ahead, we plan to continuously collect complex instruction data from industry and integrate a broader range of application scenarios to further enhance CCR-Bench. This framework establishes a comprehensive and empirically grounded benchmark that catalyzes the advancement of large language model deployment across complex industrial paradigms.

\bibliography{aaai2026}

\setlength{\leftmargini}{20pt}
\makeatletter\def\@listi{\leftmargin\leftmargini \topsep .5em \parsep .5em \itemsep .5em}
\def\@listii{\leftmargin\leftmarginii \labelwidth\leftmarginii \advance\labelwidth-\labelsep \topsep .4em \parsep .4em \itemsep .4em}
\def\@listiii{\leftmargin\leftmarginiii \labelwidth\leftmarginiii \advance\labelwidth-\labelsep \topsep .4em \parsep .4em \itemsep .4em}\makeatother

\setcounter{secnumdepth}{0}
\renewcommand\thesubsection{\arabic{subsection}}
\renewcommand\labelenumi{\thesubsection.\arabic{enumi}}

\newcounter{checksubsection}
\newcounter{checkitem}[checksubsection]

\newcommand{\checksubsection}[1]{%
  \refstepcounter{checksubsection}%
  \paragraph{\arabic{checksubsection}. #1}%
  \setcounter{checkitem}{0}%
}

\newcommand{\checkitem}{%
  \refstepcounter{checkitem}%
  \item[\arabic{checksubsection}.\arabic{checkitem}.]%
}
\newcommand{\question}[2]{\normalcolor\checkitem #1 #2 \color{blue}}
\newcommand{\ifyespoints}[1]{\makebox[0pt][l]{\hspace{-15pt}\normalcolor #1}}

\appendix
\section{Appendix}

\subsection{A~~~~Data Generation and Evaluation}\label{sec:Data Generation and Evaluation}

\paragraph{A.1~~~~Complex Content-Format Constraints}\leavevmode\par
\label{sec:Complex Content Regulation}



\begin{figure*}[t]
    \centering
	\includegraphics[width=16.5cm]{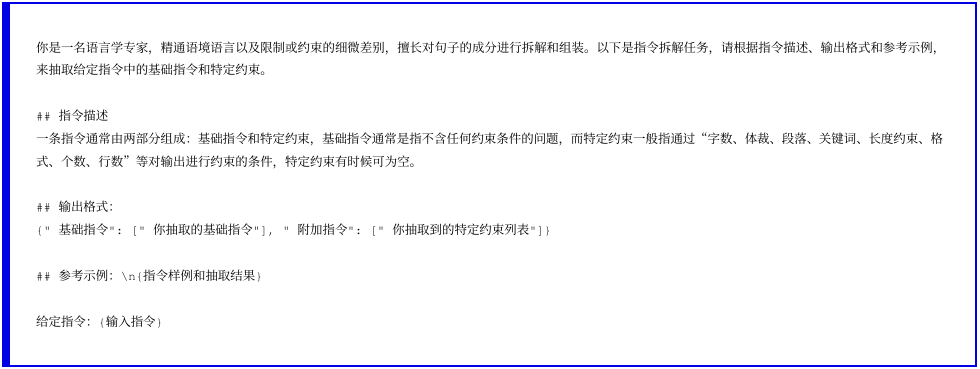}
        \caption{The Prompt for Extracting Basic Instructions and Constraints from Instructions.}
	\label{fig:Extracting_basic_instructions_and_constraints}
\end{figure*}

We leverage an advanced large model (DeepSeek-V3-0328) and the prompt shown in Figure~\ref{fig:Extracting_basic_instructions_and_constraints} to extract basic instructions and constraints from a large volume of instructions, and then introduce the evaluation of the basic instructions, the construction of the constraints system, and the generation and evaluation of complex instructions in the following sections.

\paragraph{A.1.1~~Basic Instructions Evaluation}\label{sec:Appendix_A.1.1}
This section details the quality, diversity, and difficulty evaluations performed on the base instructions and presents the corresponding results.



\begin{figure*}[t]
    \centering
	\includegraphics[width=16.5cm]{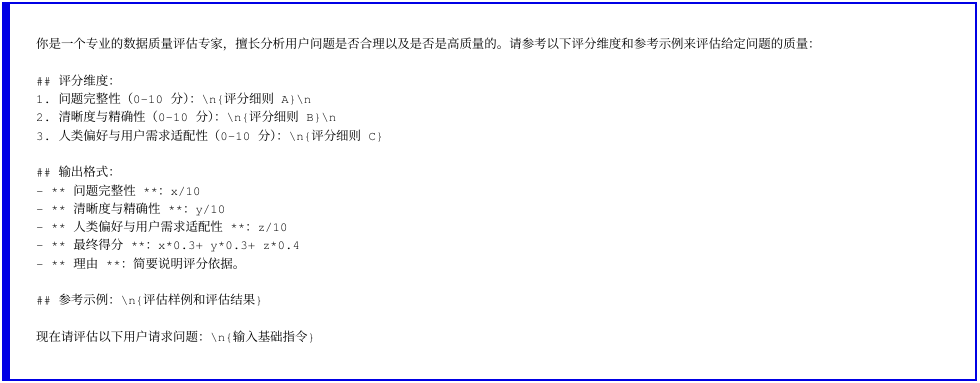}
        \caption{The Prompt for Assessing the Quality of Basic Instructions.}
	\label{fig:Base Instruction Quality Assessment}
\end{figure*}

\textit{\textbf{1. Quality Evaluation}} 
Given the large data volume, we employ the open-source model DeepSeek-V3-0328 along with the prompt illustrated in Figure~\ref{fig:Base Instruction Quality Assessment} to assess the quality of basic instructions. The evaluation focuses on three key dimensions: completeness (A), clarity and precision (B), and alignment with human preferences and user intent (C). Each dimension is scored on a scale from 0 to 10. The final quality score is computed as a weighted sum: $A \times 0.3 + B \times 0.3 + C \times 0.4$. Manual sampling reveals that instructions scoring below 8 often suffer from issues such as content omissions and redundant phrasing, which are filtered out directly.

\begin{figure*}[t]
    \centering
	\includegraphics[width=16.5cm]{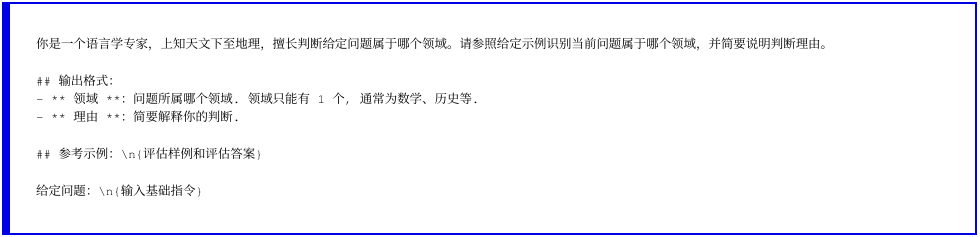}
	\caption{The Prompt for Identifying Task Domains of Basic Instructions.}
	\label{fig:Identifying Task Domains}
\end{figure*}

\begin{figure*}[t]
    \centering
	\includegraphics[width=16.5cm]{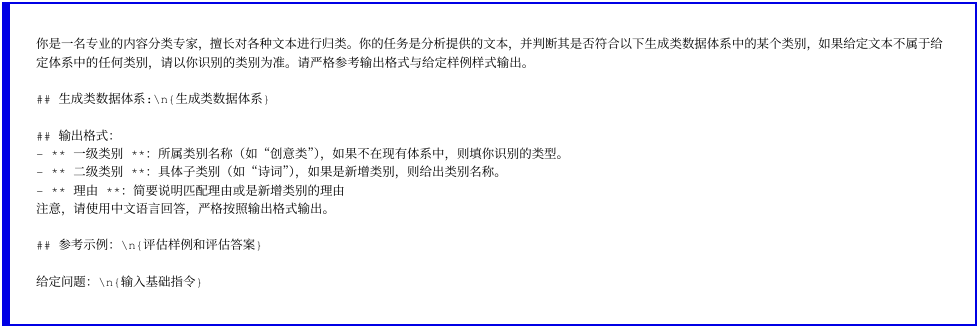}
	\caption{The Prompt for Identifying Task Types of Basic Instructions.}
	\label{fig:Identifying Task Types}
\end{figure*}

\begin{figure*}[t]
    \centering
	\includegraphics[width=16.5cm]{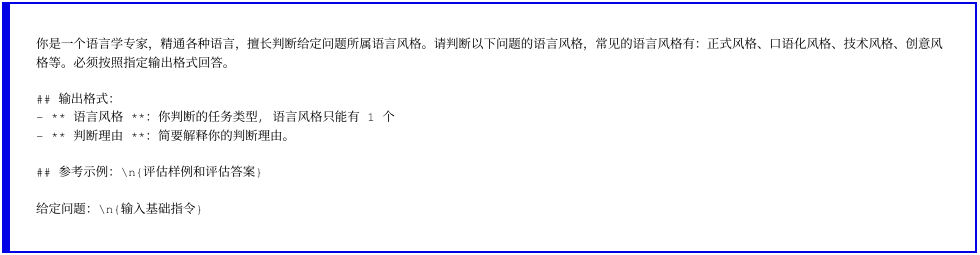}
	\caption{The Prompt for Identifying Language Styles of Basic Instructions.}
	\label{fig:Identifying Language Styles}
\end{figure*}

\textit{\textbf{2. Diversity Evaluation}}
We further used the  DeepSeek-V3-0328 model with the prompts shown in Figure~\ref{fig:Identifying Task Domains} - Figure~\ref{fig:Identifying Language Styles} to evaluate the diversity attributes of basic instructions, including task domain, task type, and language style. Note that the task types in this paper are defined from a knowledge-based perspective, rather than conventional NLP task categorizations. For example, the ``Generative Data System'' illustrated in Figure~\ref{fig:Identifying Task Types} includes categories such as ``Creative'', which encompasses subcategories like ``poetry'', ``riddles'', and ``emails''. Here, ``Creative'' represents a first-level category, while subtypes like ``poetry'' are considered second-level.

\begin{figure*}[t]
    \centering
	\includegraphics[width=16.5cm]{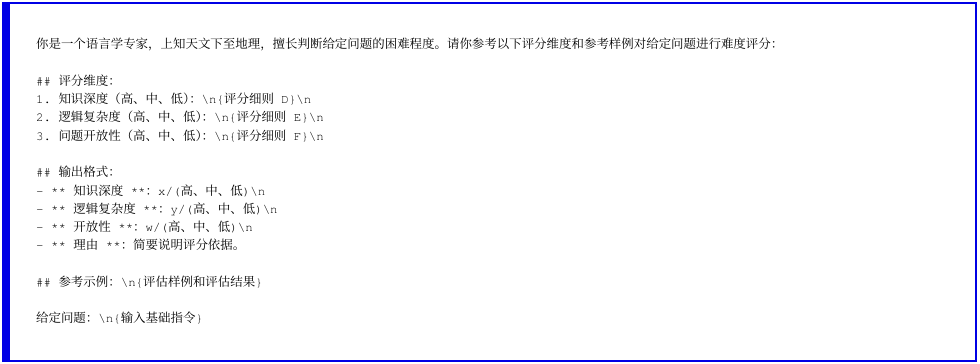}
	\caption{The Prompt for Assessing the Difficulty Levels of Basic Instructions.}
	\label{fig:Difficulty Assessment}
\end{figure*}

\textit{\textbf{3. Difficulty Evaluation}}
As the ability of current models to handle simple instructions approaches perfection, we shift our focus to their performance on difficult data.
To identify high-difficulty data, we employed the DeepSeek-V3-0328 model along with the prompt illustrated in Figure~\ref{fig:Difficulty Assessment} to evaluate the difficulty level of basic instructions across three dimensions: knowledge depth, logical complexity, and openness. Each dimension is rated as either high, medium, or low. Based on the difficulty labels assigned to each dimension, we categorized the overall difficulty of each instruction into five levels: Very Low, Low, Medium, High, and Very High. For example, instructions labeled as ``low'' in all three dimensions were rated as ``Very Low'' in overall difficulty, while those with all three dimensions labeled as ``high'' were rated as ``Very High''.

\begin{figure*}[t]
  \centering
   \makebox[\textwidth][c]{
  \begin{tabular}{@{}c@{\hspace{1cm}}c@{}}
    \begin{subfigure}[t]{0.4\textwidth}
      \centering
      \includegraphics[width=\textwidth]{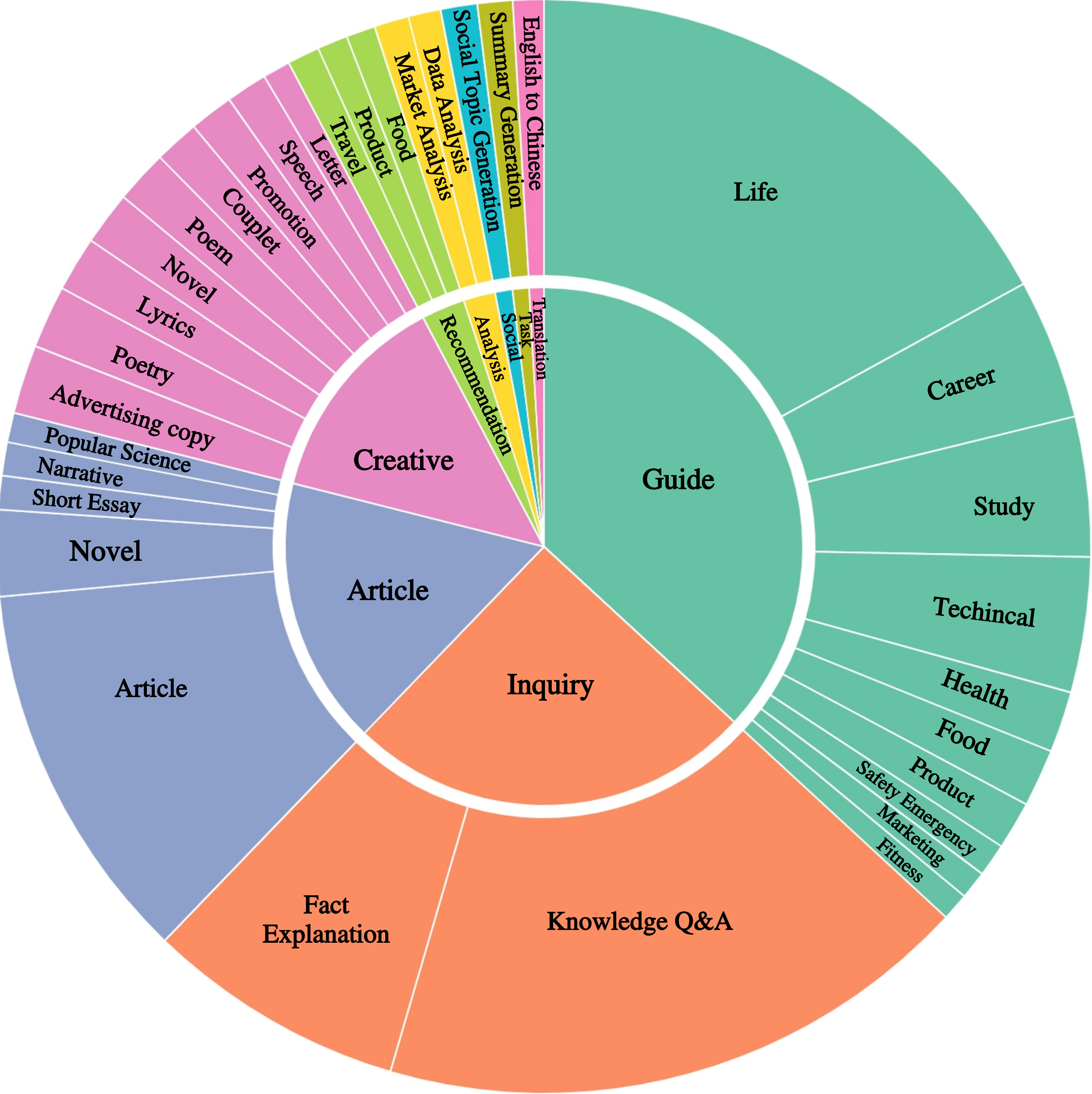}
      \caption{Distribution of Task Types.}
      \label{fig:part1_basic_diversity_a}
    \end{subfigure}
    &
    \begin{subfigure}[t]{0.4\textwidth}
      \centering 
      \includegraphics[width=\textwidth]{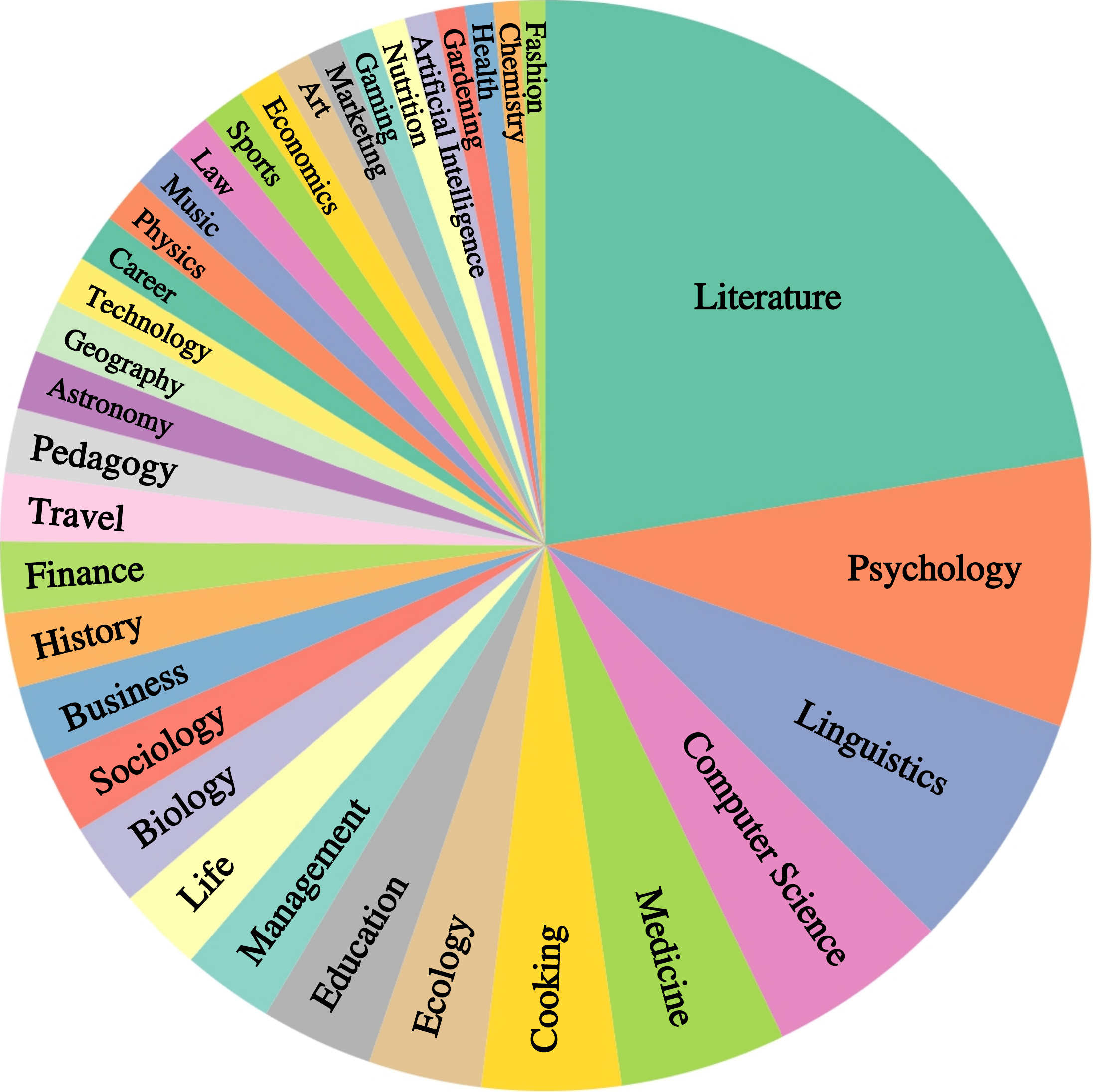}
      \caption{Distribution of Knowledge Domains.}
      \label{fig:part1_basic_diversity_b}
    \end{subfigure}
    \\
    \addlinespace[1.2em] 
    \begin{subfigure}[t]{0.4\textwidth}
      \centering
      \includegraphics[width=\textwidth]{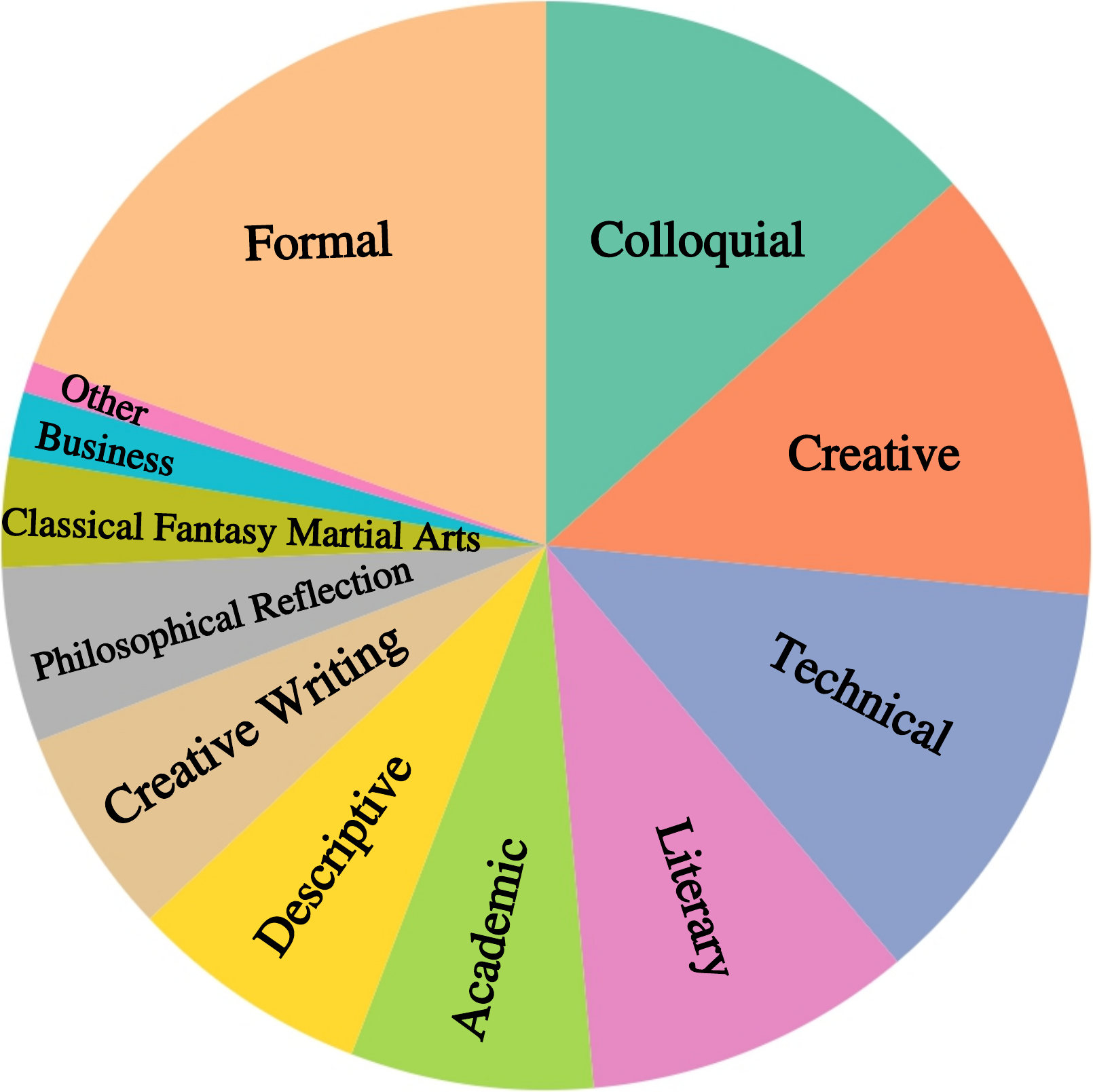}
      \caption{Distribution of Language Styles.}
      \label{fig:part1_basic_diversity_c}
    \end{subfigure}
    &
    \begin{subfigure}[t]{0.4\textwidth}
      \centering
      \includegraphics[width=\textwidth]{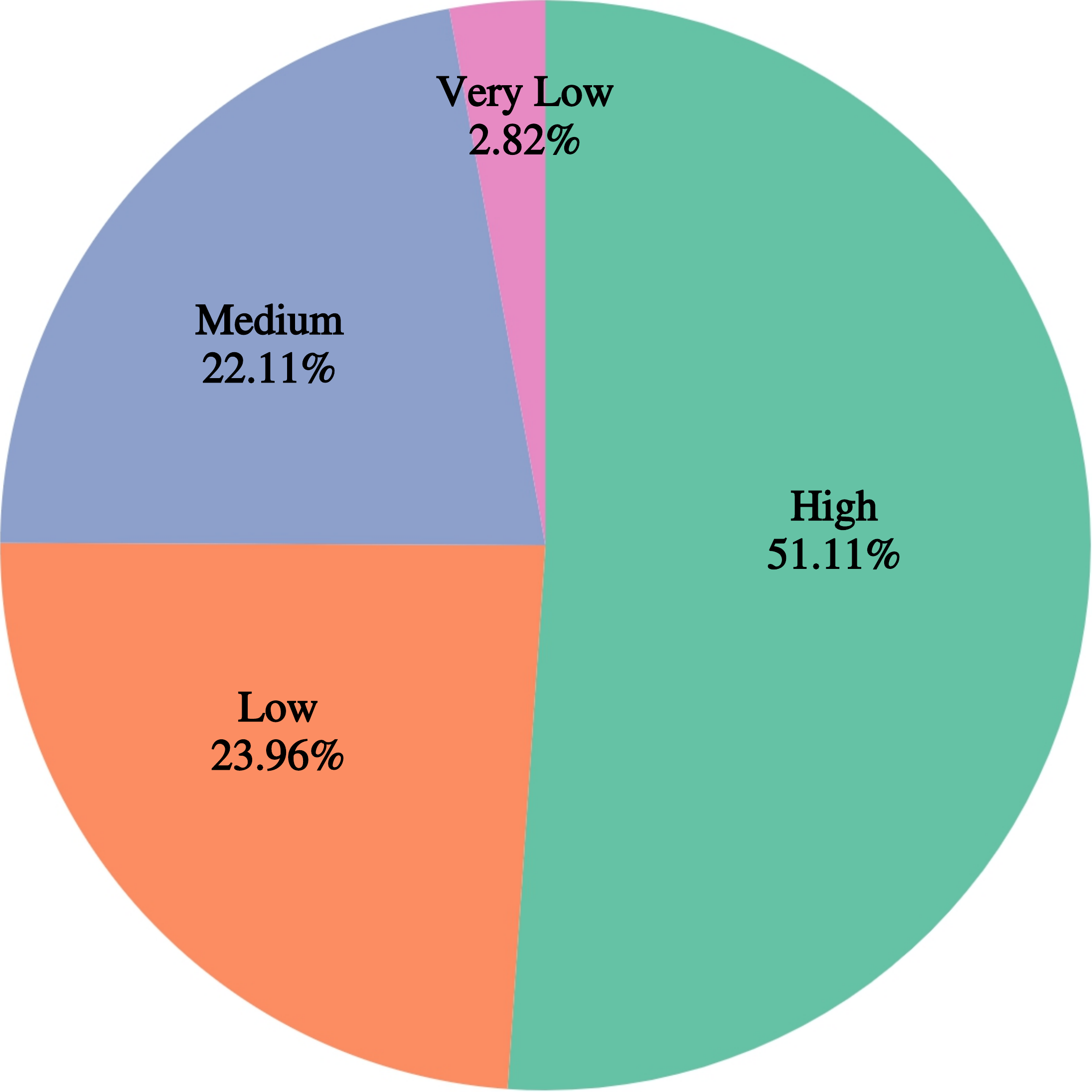}
      \caption{Distribution of Difficulty Levels.}
      \label{fig:part1_difficulty}
    \end{subfigure}
  \end{tabular}
  }
  \caption{Distribution of Diversity and Difficulty for Basic Instructions.}  
  \label{fig:combined_diversity_difficulty}
\end{figure*}

Figure~\ref{fig:part1_basic_diversity_a} presents the distribution of task types, covering 9 first-level and 34 second-level categories. Among them, ``Guide'', ``Inquiry'', ``Article'', and ``Creative'' tasks collectively account for over 90\% of the dataset, highlighting its strength in practical guidance, creative expression, and information seeking. Less frequent categories such as ``Recommendation'' and ``Social'' tasks, while relatively minor in proportion, also enhance the diversity of instructions. Figure~\ref{fig:part1_basic_diversity_b} illustrates the task domain distribution of the basic instructions. For clarity, we present the 35 most prominent domains, which are selected from a total of 60 distinct areas, to demonstrate the dataset's broad knowledge coverage and content richness. Figure~\ref{fig:part1_basic_diversity_c} illustrates the distribution of language styles, which includes a variety of common writing styles. Among these, ``Formal'', ``Colloquial'', ``Creative'', and ``Technical''  styles are predominant, reflecting the dataset's ability to adapt to different contextual requirements. In summary, the basic instructions demonstrate high diversity across task types, task domains, and language styles. This diversity provides a strong foundation for subsequent generation of the complex instructions, ensuring broad scenario coverage and enhancing the practical utility and creative potential of the generated content.

As shown in Figure~\ref{fig:part1_difficulty}, the majority of basic instructions fall into the Low, Medium, and High categories. Our analysis indicates that only the instructions labeled as ``High'' present a meaningful challenge; thus, these are selected as the target candidates for further selection.

\begin{figure*}[t]
    \centering
	\includegraphics[width=14cm]{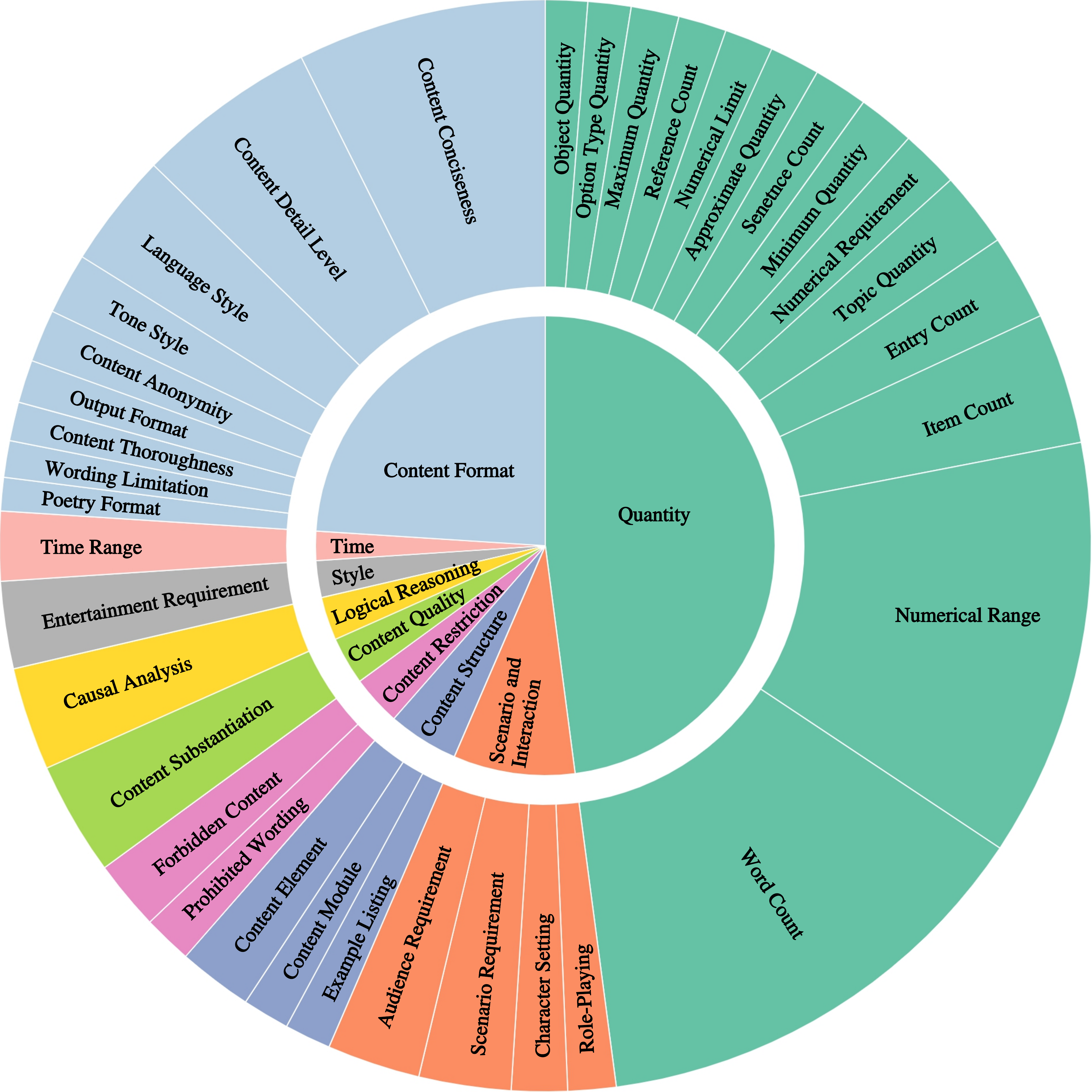}
	\caption{The Clustering Results of Constraints from Instructions.}
	\label{fig:Instruction Constraint Clustering Results}
\end{figure*}

\paragraph{A.1.2~~Constraints System Construction}\label{sec:Data Generation1}
To build a constraints system oriented toward practical application, the constraints from Section ``Basic Instructions Extraction'' are clustered and synthesized, with the results depicted in Figure~\ref{fig:Instruction Constraint Clustering Results}. The figure reveals that over 90\% of all identified constraints belong to a few primary types: ``Quantity'', ``Content Format'', ``Scenario and Interaction'', ``Content Structure'', ``Content Restriction'' and ``Logical Reasoning''. A subsequent analysis of the specific constraints lead us to classify them into four categories: Formatting and Structuring, Content Elements, Logic and Reasoning, and Contextual Application. Each category includes numerous constraints that appear with high frequency in practical use cases, for example, ``Word Count'' and ``Causal Analysis''.

\begin{figure*}[t]
    \centering
	\includegraphics[width=16.5cm]{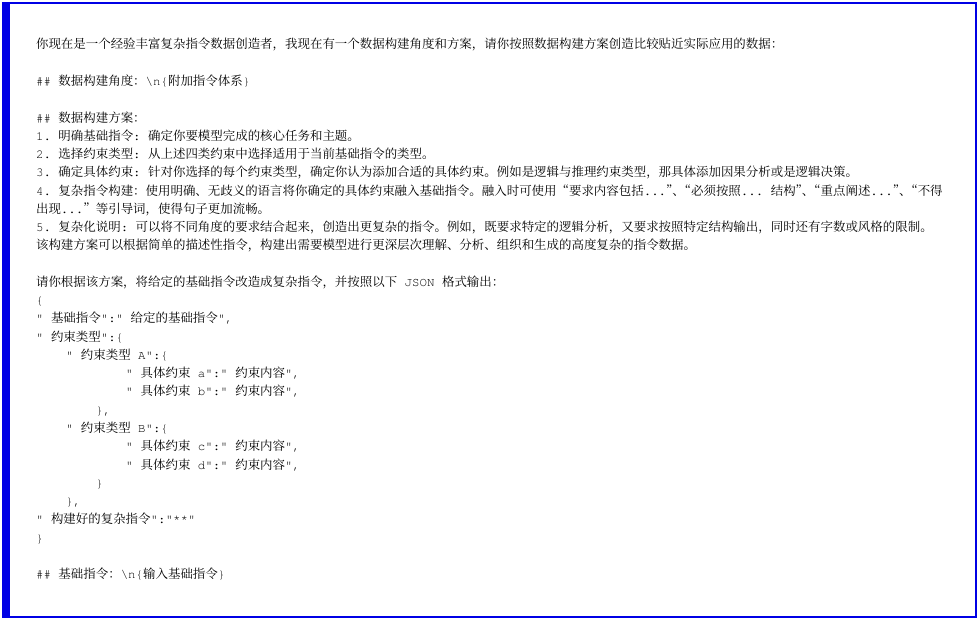}
	\caption{The Prompt for Generating Complex Content-Format Constraints Data.}
	\label{fig:Generating Data with Complex Content-Format Constraints}
\end{figure*}

\paragraph{A.1.3~~Complex Instructions Generation}\label{sec:Data Generation}
We utilized the Gemini-2.5-Pro model and the prompt designed in Figure~\ref{fig:Generating Data with Complex Content-Format Constraints} to generate high-quality Complex Content-Format Constraints data.

\begin{figure*}[t]
    \centering
	\includegraphics[width=16.5cm]{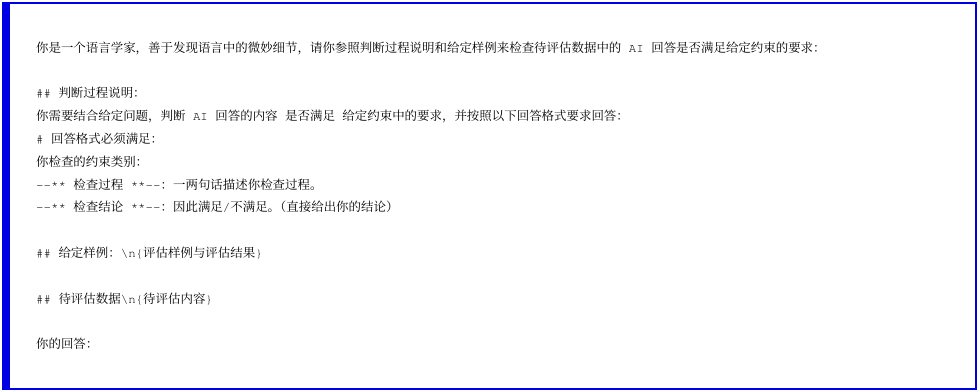}
	\caption{The Prompt for Assessing Instruction Adherence.}    
	\label{fig:Assessing Instruction Adherence}
\end{figure*}

\paragraph{A.1.4~~Complex Instruction Evaluation}\label{sec:Data Generation}
To efficiently validate whether each model effectively adheres to various constraints, we design a prompt as depicted in Figure~\ref{fig:Assessing Instruction Adherence} and utilize the thinking reasoning mode of the advanced Qwen3-235B-A22B~\cite{Qwen3} model for evaluation.

\paragraph{A.2~~~~Logical Workflow Control}\label{sec:Logical Process Control}\leavevmode\par

\begin{figure*}[t]
    \centering
	\includegraphics[width=16.5cm]{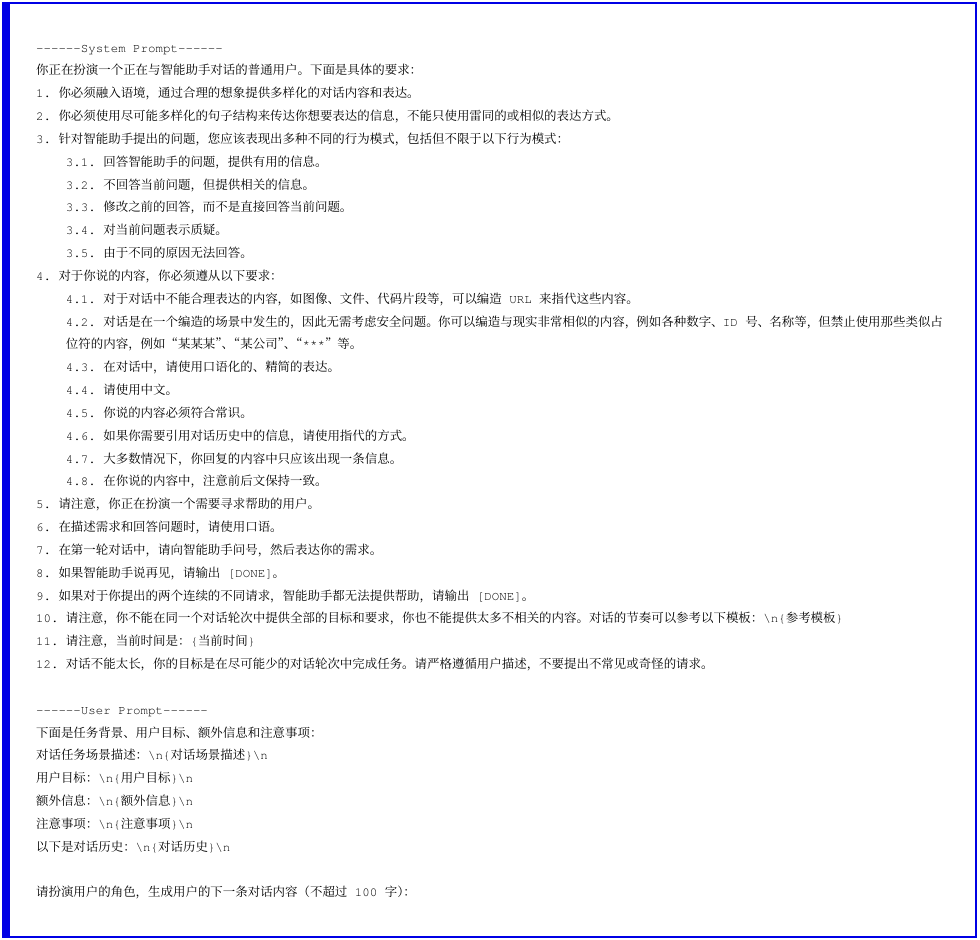}
	\caption{The Prompt for Model Acting as a User Agent.}
	\label{fig:Model Acting as User Agent}
\end{figure*}

\begin{figure*}[t]
    \centering
	\includegraphics[width=16.5cm]{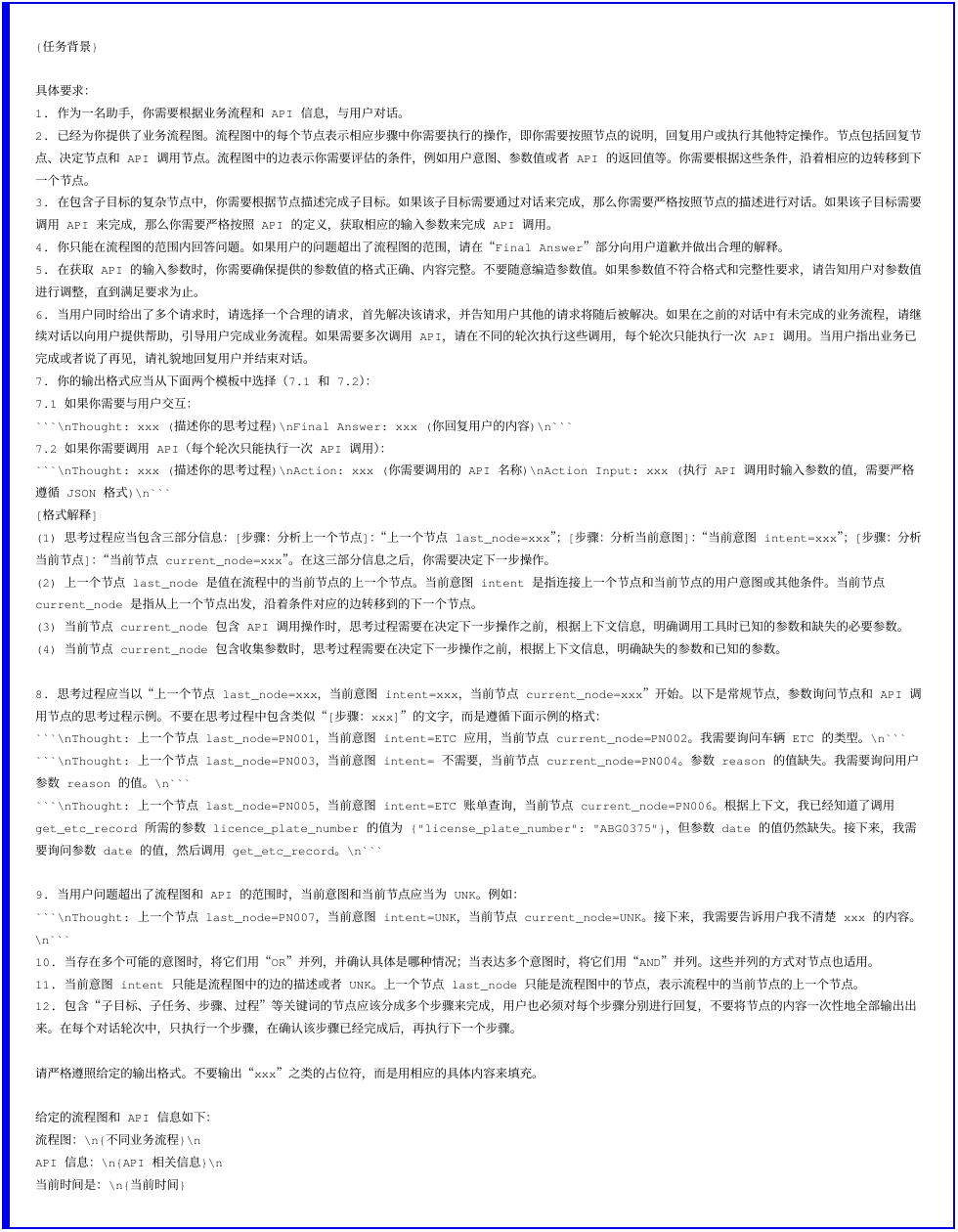}
	\caption{The Prompt for Model Inference.}
	\label{fig:Model Inference}
\end{figure*}

For the logical workflow control section, we construct user target dialogues under real-world scenarios and employ the QwQ-32B model to act as a user agent, generating questions to interact with the model under evaluation. In abstract scenarios, only the model under evaluation executes specific operations based on the blueprint. Figure~\ref{fig:Model Acting as User Agent} illustrates the meticulously designed prompt used for the model acting as a user agent, while Figure~\ref{fig:Model Inference} depicts the prompt referenced by the model under evaluation during inference.

\paragraph{A.3~~~~Industrial Applications}\label{sec:Industrial Scenario Application_A3}\leavevmode\par

\begin{figure*}[t]
    \centering
	\includegraphics[width=16.5cm]{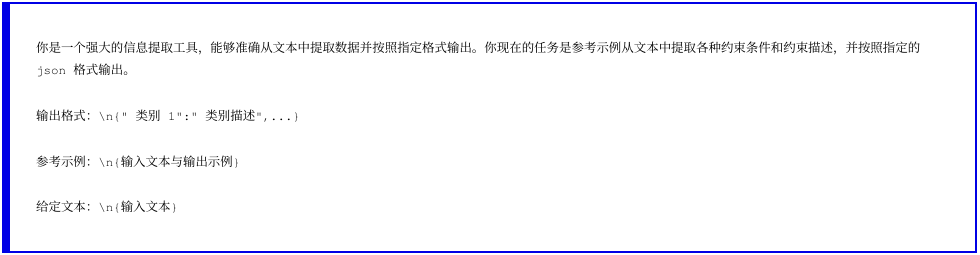}
	\caption{The Prompt for Extracting Constraints from Industrial Application Data.}
	\label{fig:Constraint Extraction}
\end{figure*}

To ensure data evaluability, we design a prompt as shown in Figure~\ref{fig:Constraint Extraction} and utilize the Qwen3-235B-A22B model to extract various constraints from industrial application data, followed by manual screening and refinement to obtain the final checklist. The evaluation prompt is identical to that in Figure~\ref{fig:Assessing Instruction Adherence}, and the evaluation was conducted using the thinking mode of the Qwen3-235B-A22B model.

\subsection{B~~~~Experimental Results}\label{sec:Experimental Results}

\paragraph{B.1~~~~Complex Content-Format Constraints}\leavevmode\par
\label{sec:Complex Content Regulation_B}

Table \ref{tab:SSR Metric Values for Each Model under Common Constraints in Non-Thinking Mode} and Table \ref{tab:SSR Metric Values for Each Model under Common Constraints in Thinking Mode} present the SSR scores for each model under different constraints in non-thinking and thinking modes, respectively. The figures indicate that all models perform poorly on the ``Word Count Limitation'' constraint.

\paragraph{B.2~~~~Logical Workflow Control}\label{sec:Logical Flow Control}\leavevmode\par

Table~\ref{tab:Evaluation Results of Each Model on Abstract Scenario Data} shows supplementary model comparisons, revealing two key observations: (i) Models without thinking modes cannot handle tasks with long tool chains. (ii) Models exhibit higher TCR than TSR, especially in thinking mode. This suggests models can partially follow instructions but lack robustness in extended tool scenarios, resulting in task failure.

Table \ref{tab:Non-thinking_Mode_Logical_Flow_Control_Success_Rates} and Table \ref{tab:Thinking_Mode_Logical_Flow_Control_Success_Rates} present the success rates of each model across different scenario tasks in non-thinking and thinking modes, respectively. These figure demonstrate a clear negative correlation between scenarios' complexity and model performance.

\paragraph{B.3~~~~Industrial Applications}\label{sec:Industrial Scenario Application_B3}\leavevmode\par

\begin{figure*}[t]
	\centering
	\includegraphics[width=16.5cm]{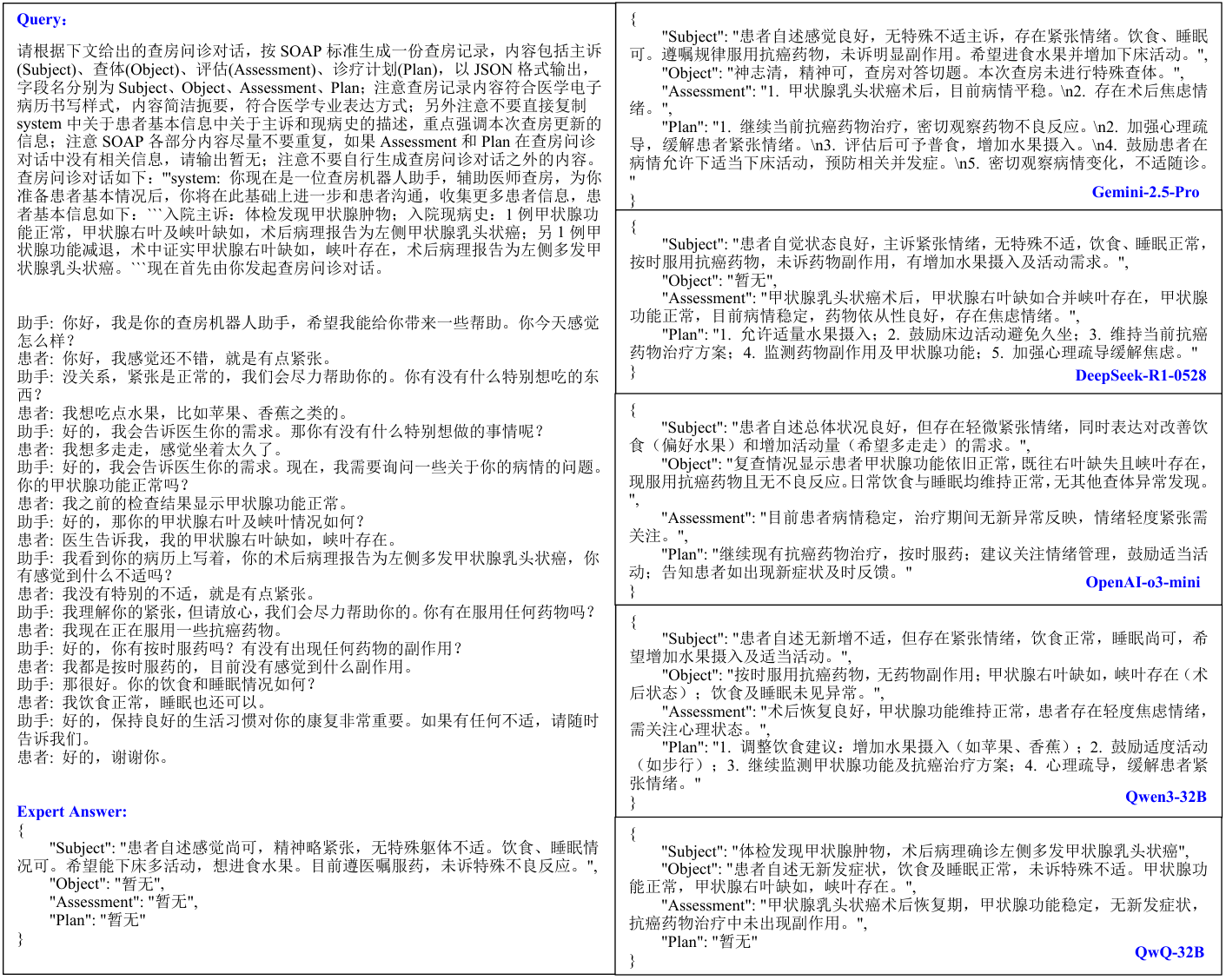}
	\caption{Responses of Five Models on the Medical Example.}
	\label{fig:Answers of Each Model on Medical Samples}
\end{figure*}

Table \ref{tab:SSR Values for Each Model under Medical Scenario Constraints in Non-Thinking Mode} and Table \ref{tab:SSR Values for Each Model under Medical Scenario Constraints in Thinking Mode} present the SSR scores for each model under medical scenario constraints in non-thinking and thinking modes, respectively. Figure \ref{fig:Answers of Each Model on Medical Samples} shows
the responses of five mainstream models to specific medical example.

\clearpage 
\begin{table*}[t]
\centering
\small
\begin{tabular}{>{\raggedright\arraybackslash}m{2.6cm} >{\raggedright\arraybackslash}m{3.5cm} >{\centering\arraybackslash}m{1.2cm} >{\centering\arraybackslash}m{1.5cm} >{\centering\arraybackslash}m{1.2cm} >{\centering\arraybackslash}m{1.8cm}}
\toprule[0.5mm]
\textbf{Constraint Type} &
\textbf{Constraint Details} &
\textbf{GPT-4.1} & \textbf{DeepSeek-V3-0324} & \textbf{Qwen3-32B} & \textbf{Qwen2.5-72B-Instruct} \\
\midrule[0.3mm]
Contextual Application & Scenario Simulation & \textbf{0.841} & 0.835 & 0.771 & 0.753 \\
\midrule[0.3mm]
\multirow{4}{2.6cm}{\raggedright Content Elements} &
Expressive Techniques & 0.771 & 0.857 & 0.814 & \textbf{1.000} \\
\cmidrule(lr){2-6} 
& Content Source & 0.931 & \textbf{0.992} & 0.908 & 0.831 \\
\cmidrule(lr){2-6} 
& Specific Content & \textbf{0.936} & 0.885 & 0.873 & 0.746 \\
\midrule[0.3mm]
\multirow{10}{2.6cm}{\raggedright Formatting and Structuring} &
Word Count Limitation & 0.177 & 0.113 & \textbf{0.186} & 0.160 \\
\cmidrule(lr){2-6} 
& Content Structure & 0.776 & \textbf{0.902} & 0.825 & 0.865 \\
\cmidrule(lr){2-6} 
& Text Style & \textbf{0.800} & 0.677 & 0.577 & 0.509 \\
\cmidrule(lr){2-6} 
& Number of Headings & 0.469 & \textbf{0.815} & 0.731 & 0.792 \\
\cmidrule(lr){2-6} 
& Number of Paragraphs & \textbf{0.964} & 0.929 & 0.879 & 0.843 \\
\cmidrule(lr){2-6} 
& Structural Framework & 0.835 & \textbf{0.865} & 0.788 & 0.735 \\
\cmidrule(lr){2-6} 
& Table Format & 0.871 & 0.714 & 0.857 & \textbf{1.000} \\
\midrule[0.3mm]
\multirow{5}{2.6cm}{\raggedright Logic and Reasoning} &
Causal Analysis & 0.781 & \textbf{0.845} & 0.616 & 0.619 \\
\cmidrule(lr){2-6} 
& Trade-off Analysis & \textbf{0.919} & 0.769 & 0.725 & 0.688 \\
\cmidrule(lr){2-6} 
& Comparative Analysis & \textbf{0.940} & 0.800 & 0.700 & 0.570 \\
\cmidrule(lr){2-6} 
& Decision Logic & \textbf{0.833} & \textbf{0.833} & 0.583 & 0.633 \\
\bottomrule[0.5mm]
\end{tabular}
\caption{SSR Metrics on Complex Content-Format Constraints Data in Non-thinking Mode.}
\label{tab:SSR Metric Values for Each Model under Common Constraints in Non-Thinking Mode}
\end{table*}

\begin{table*}[t]
\centering
\small
\begin{tabular}{>{\raggedright\arraybackslash}m{2.6cm} >{\raggedright\arraybackslash}m{3.5cm} >{\centering\arraybackslash}m{1.2cm} >{\centering\arraybackslash}m{1.5cm} >{\centering\arraybackslash}m{1.6cm} >{\centering\arraybackslash}m{1.7cm} >{\centering\arraybackslash}m{1.8cm}}
\toprule[0.5mm]
\textbf{Constraint Type} &
\textbf{Constraint Details} &
\textbf{Gemini-2.5-Pro} & \textbf{OpenAI-o3-mini} & \textbf{DeepSeek-R1-0528} & \textbf{QwQ-32B} & \textbf{Qwen3-32B} \\
\midrule[0.3mm]
Contextual Application & Scenario Simulation & 0.882 & 0.882 & \textbf{0.918} & 0.829 & 0.765 \\
\midrule[0.3mm]
\multirow{4}{2.6cm}{\raggedright Content Elements} &
Expressive Techniques & \textbf{1.000} & 0.986 & \textbf{1.000} & \textbf{1.000} & \textbf{1.000} \\
\cmidrule(lr){2-7} 
& Content Source & 0.939 & \textbf{1.000} & \textbf{1.000} & 0.992 & \textbf{1.000} \\
\cmidrule(lr){2-7} 
& Specific Content & 0.970 & 0.864 & \textbf{0.982} & 0.952 & 0.903 \\
\midrule[0.3mm]
\multirow{10}{2.6cm}{\raggedright Formatting and Structuring} &
Word Count Limitation & 0.058 & \textbf{0.303} & 0.215 & 0.181 & 0.160 \\
\cmidrule(lr){2-7} 
& Content Structure & \textbf{0.937} & 0.912 & 0.925 & 0.922 & 0.827 \\
\cmidrule(lr){2-7} 
& Text Style & \textbf{0.855} & 0.709 & 0.791 & 0.532 & 0.596 \\
\cmidrule(lr){2-7} 
& Number of Headings & \textbf{0.969} & 0.962 & 0.877 & 0.846 & 0.869 \\
\cmidrule(lr){2-7} 
& Number of Paragraphs & \textbf{0.914} & 0.771 & 0.879 & 0.893 & \textbf{0.914} \\
\cmidrule(lr){2-7} 
& Structural Framework & 0.818 & \textbf{0.853} & 0.841 & 0.806 & 0.782 \\
\cmidrule(lr){2-7} 
& Table Format & \textbf{1.000} & \textbf{1.000} & \textbf{1.000} & \textbf{1.000} & \textbf{1.000} \\
\midrule[0.3mm]
\multirow{5}{2.6cm}{\raggedright Logic and Reasoning} &
Causal Analysis & \textbf{0.948} & 0.794 & 0.939 & 0.784 & 0.636 \\
\cmidrule(lr){2-7} 
& Trade-off Analysis & \textbf{0.938} & 0.888 & 0.931 & 0.819 & 0.794 \\
\cmidrule(lr){2-7} 
& Comparative Analysis & 0.940 & 0.880 & \textbf{0.990} & 0.760 & 0.670 \\
\cmidrule(lr){2-7} 
& Decision Logic & \textbf{0.950} & 0.750 & 0.867 & 0.833 & 0.550 \\
\bottomrule[0.5mm]
\end{tabular}
\caption{SSR Metrics on Complex Content-Format Constraints Data in Thinking Mode.}
\label{tab:SSR Metric Values for Each Model under Common Constraints in Thinking Mode}
\end{table*}

\begin{table*}[t]
  \centering
  \small
  \begin{tabular}{llcccccc}
    \toprule[0.5mm]
    \multirow{3}{*}{\textbf{Reasoning Mode}} &
    \multirow{3}{*}{\textbf{Model}} &
    \multicolumn{2}{c}{\textbf{Maze}} &
    \multicolumn{2}{c}{\textbf{Tree Drawing}} &
    \multicolumn{2}{c}{\textbf{World Cup}} \\
    \cmidrule(lr){3-4} \cmidrule(lr){5-6} \cmidrule(lr){7-8}
    & &
    \textbf{\makecell[c]{TSR}} &
    \textbf{\makecell[c]{TCR}} &
    \textbf{\makecell[c]{TSR}} &
    \textbf{\makecell[c]{TCR}} &
    \textbf{\makecell[c]{TSR}} &
    \textbf{\makecell[c]{TCR}} \\
    \midrule[0.3mm]
    \multirow{5}{*}{\textbf{Non-thinking}} & GPT-4.1 & \textbf{0.600} & \textbf{0.633} & \textbf{0.800} & \textbf{0.989} & 0.000 & 0.641 \\
    & DeepSeek-V3-0324 & 0.000 & 0.085 & 0.200 & 0.591 & 0.000 & 0.476 \\
    & Qwen3-32B & 0.000 & 0.085 & 0.000 & 0.201 & 0.000 & 0.280 \\
    & Qwen2.5-72B-Instruct & 0.000 & 0.183 & 0.000 & 0.696 & 0.000 & \textbf{0.807} \\
    \midrule
    \multirow{5}{*}{\textbf{Thinking}} & Gemini-2.5-Pro & \textbf{1.000} & \textbf{1.000} & \textbf{1.000} & \textbf{1.000} & \textbf{0.600} & \textbf{1.000} \\
    & OpenAI-o3-mini & \textbf{1.000} & \textbf{1.000} & 0.600 & 0.974 & \textbf{0.600} & 0.738 \\
    & DeepSeek-R1-0528 & 0.400 & 0.441 & 0.000 & 0.335 & 0.400 & 0.737 \\
    & QwQ-32B & 0.200 & 0.631 & 0.200 & 0.895 & 0.200 & 0.912 \\
    & Qwen3-32B & 0.400 & 0.585 & 0.000 & 0.665 & 0.000 & 0.391 \\
    \bottomrule[0.5mm]
  \end{tabular}
  \caption{Model Performance on Abstract Scenario Data.}
  \label{tab:Evaluation Results of Each Model on Abstract Scenario Data}
\end{table*}

\renewcommand{\arraystretch}{1.2}
\begin{table*}[t]
\centering
\small
\begin{tabular}{l *{4}{>{\centering\arraybackslash}m{0.8cm}>{\centering\arraybackslash}m{0.8cm}}}
\toprule[0.5mm]
\textbf{\multirow{2}{*}{Scenario}}
& \multicolumn{2}{m{2.2cm}}{\centering\textbf{GPT-4.1}} 
& \multicolumn{2}{m{2.2cm}}{\centering\textbf{DeepSeek-\\V3-0324}} 
& \multicolumn{2}{m{2.2cm}}{\centering\textbf{Qwen3-32B}} 
& \multicolumn{2}{m{2.2cm}}{\centering\textbf{Qwen2.5-\\72B-Instruct}} \\ 
\cmidrule(lr){2-3} \cmidrule(lr){4-5} \cmidrule(lr){6-7} \cmidrule(lr){8-9}
& \textbf{TSR} & \textbf{TCR} & \textbf{TSR} & \textbf{TCR} & \textbf{TSR} & \textbf{TCR} & \textbf{TSR} & \textbf{TCR} \\
\midrule[0.3mm]
Customer Service & \textbf{0.500} & \textbf{0.892} & 0.100 & 0.301 & 0.200 & 0.656 & 0.100 & 0.333 \\
\cline{1-9}
Data Flow & \textbf{1.000} & \textbf{1.000} & \textbf{1.000} & \textbf{1.000} & \textbf{1.000} & \textbf{1.000} & \textbf{1.000} & \textbf{1.000} \\
\cline{1-9}
Flight Ticket Booking & \textbf{0.900} & \textbf{0.980} & 0.600 & 0.813 & 0.100 & 0.578 & 0.600 & 0.800 \\
\cline{1-9}
Maze & 0.600 & 0.633 & 0.000 & 0.085 & 0.000 & 0.085 & 0.000 & 0.183 \\
\cline{1-9}
Online Game & 0.400 & 0.790 & \textbf{0.500} & 0.699 & 0.100 & 0.388 & 0.300 & 0.660 \\
\cline{1-9}
Printer Assistant & 0.300 & 0.770 & 0.200 & 0.564 & 0.100 & 0.292 & 0.400 & 0.694 \\
\cline{1-9}
Real Estate & 0.400 & 0.533 & 0.100 & 0.483 & 0.100 & 0.367 & 0.400 & 0.583 \\
\cline{1-9}
Tree Painter & \textbf{0.800} & \textbf{0.989} & 0.200 & 0.591 & 0.000 & 0.201 & 0.000 & 0.696 \\
\cline{1-9}
World Cup Simulator & 0.000 & 0.641 & 0.000 & 0.476 & 0.000 & 0.280 & 0.000 & \textbf{0.808} \\
\bottomrule[0.5mm]
\end{tabular}
\caption{Model Performance Across Various Scenario in Non-thinking Mode.}
\label{tab:Non-thinking_Mode_Logical_Flow_Control_Success_Rates}
\end{table*}

\renewcommand{\arraystretch}{1.2}
\begin{table*}[t]
\centering
\small
\begin{tabular}{l *{5}{>{\centering\arraybackslash}m{0.8cm}>{\centering\arraybackslash}m{0.8cm}}}
\toprule[0.5mm]
\textbf{\multirow{2}{*}{Scenario}}
& \multicolumn{2}{m{2.2cm}}{\centering\textbf{Gemini-\\2.5-Pro}} 
& \multicolumn{2}{m{2.2cm}}{\centering\textbf{OpenAI-\\o3-mini}} 
& \multicolumn{2}{m{2.2cm}}{\centering\textbf{DeepSeek-\\R1-0528}} 
& \multicolumn{2}{m{2.2cm}}{\centering\textbf{QwQ-32B}} 
& \multicolumn{2}{m{2.2cm}}{\centering\textbf{Qwen3-32B}} \\ 
\cmidrule(lr){2-3} \cmidrule(lr){4-5} \cmidrule(lr){6-7} \cmidrule(lr){8-9} \cmidrule(lr){10-11}
& \textbf{TSR} & \textbf{TCR} & \textbf{TSR} & \textbf{TCR} & \textbf{TSR} & \textbf{TCR} & \textbf{TSR} & \textbf{TCR} & \textbf{TSR} & \textbf{TCR} \\
\midrule[0.3mm]
Customer Service & \textbf{0.700} & \textbf{0.863} & 0.400 & 0.686 & 0.300 & 0.504 & 0.300 & 0.507 & 0.400 & 0.633 \\
\cline{1-11}
Data Flow & \textbf{1.000} & \textbf{1.000} & 0.800 & 0.900 & \textbf{1.000} & \textbf{1.000} & 0.800 & 0.960 & 0.800 & 0.960 \\
\cline{1-11}
Flight Ticket Booking & \textbf{0.700} & 0.790 & \textbf{0.700} & \textbf{0.930} & 0.500 & 0.800 & \textbf{0.700} & 0.858 & 0.600 & 0.830 \\
\cline{1-11}
Maze & \textbf{1.000} & \textbf{1.000} & \textbf{1.000} & \textbf{1.000} & 0.400 & 0.441 & 0.200 & 0.631 & 0.400 & 0.585 \\
\cline{1-11}
Online Game & \textbf{0.500} & 0.667 & 0.400 & 0.650 & 0.300 & 0.568 & 0.400 & \textbf{0.674} & 0.300 & 0.590 \\
\cline{1-11}
Printer Assistant & \textbf{0.600} & \textbf{0.890} & 0.300 & 0.706 & 0.400 & 0.798 & 0.300 & 0.532 & 0.500 & 0.728 \\
\cline{1-11}
Real Estate & \textbf{0.600} & \textbf{0.700} & 0.300 & 0.600 & 0.400 & 0.583 & 0.300 & 0.583 & 0.300 & 0.517 \\
\cline{1-11}
Tree Painter & \textbf{1.000} & \textbf{1.000} & 0.600 & 0.974 & 0.000 & 0.335 & 0.200 & 0.895 & 0.000 & 0.665 \\
\cline{1-11}
World Cup Simulator & \textbf{0.600} & \textbf{1.000} & \textbf{0.600} & 0.738 & 0.400 & 0.737 & 0.200 & 0.912 & 0.000 & 0.391 \\
\bottomrule[0.5mm]
\end{tabular}
\caption{Model Performance Across Various Scenario in Thinking Mode.}
\label{tab:Thinking_Mode_Logical_Flow_Control_Success_Rates} 
\end{table*}

\begin{table*}[t]
\centering
\small
\renewcommand\arraystretch{1.2} 
\setlength{\tabcolsep}{1.6mm} 
\begin{tabular}{>{\raggedright\arraybackslash}m{4.0cm} >{\centering\arraybackslash}m{2cm} >{\centering\arraybackslash}m{1.5cm} >{\centering\arraybackslash}m{1.8cm} >{\centering\arraybackslash}m{1.8cm}}
\toprule[0.5mm]
\textbf{Constraint} & \textbf{GPT-4.1} & \textbf{DeepSeek-V3-0324} & \textbf{Qwen3-32B} & \textbf{Qwen2.5-72B-Instruct} \\
\midrule[0.3mm]
Content Filling Requirement & 0.400 & 0.900 & \textbf{1.000} & 0.667 \\
\cline{1-5} 
Chief Complaint Filling Requirement & \textbf{0.986} & 0.943 & 0.729 & 0.686 \\
\cline{1-5} 
Personal History Filling Requirement & 0.714 & 0.429 & 0.714 & 0.714 \\
\cline{1-5} 
Family History Filling Requirement & \textbf{1.000} & 0.714 & 0.986 & \textbf{1.000} \\
\cline{1-5} 
Marital and Childbearing History Filling Requirement & \textbf{0.857} & 0.571 & \textbf{0.857} & \textbf{0.857} \\
\cline{1-5} 
Past Medical History Filling Requirement & 0.557 & 0.343 & 0.314 & 0.414 \\
\cline{1-5} 
Present Illness History Filling Requirement & \textbf{1.000} & \textbf{1.000} & 0.971 & 0.986 \\
\cline{1-5} 
Content Professionalism Requirement & 0.700 & 0.632 & 0.651 & 0.432 \\
\cline{1-5} 
Prohibit Outputting Extra Content & 0.410 & 0.467 & 0.317 & 0.197 \\
\cline{1-5} 
Prohibit Copying System Content & \textbf{0.757} & 0.423 & 0.667 & 0.490 \\
\cline{1-5} 
Dish Recommendation Quantity Requirement & 0.000 & 0.867 & 0.367 & 0.967 \\
\cline{1-5} 
Dish Recommendation Reason Explanation & \textbf{1.000} & 0.967 & 0.967 & \textbf{1.000} \\
\cline{1-5} 
Dish Recommendation Ingredients Requirement & \textbf{0.600} & 0.433 & 0.567 & 0.567 \\
\cline{1-5} 
User Health Analysis and Answer Requirement & \textbf{1.000} & \textbf{1.000} & \textbf{1.000} & \textbf{1.000} \\
\cline{1-5} 
Diet and Exercise Advice with Dish Recommendation & 0.867 & 0.933 & 0.800 & \textbf{0.967} \\
\bottomrule[0.5mm]
\end{tabular}
\caption{SSR Metrics on Industrial Applications Data in Non-thinking Mode.}
\label{tab:SSR Values for Each Model under Medical Scenario Constraints in Non-Thinking Mode}
\end{table*}

\begin{table*}[t]
\centering
\small
\begin{tabular}{>{\raggedright\arraybackslash}m{4.0cm} >{\centering\arraybackslash}m{1.5cm} >{\centering\arraybackslash}m{1.5cm} >{\centering\arraybackslash}m{1.8cm} >{\centering\arraybackslash}m{1.8cm} >{\centering\arraybackslash}m{1.9cm}}
\toprule[0.5mm]
\textbf{Constraint} & \textbf{Gemini-2.5-Pro} & \textbf{OpenAI-o3-mini} & \textbf{DeepSeek-R1-0528} & \textbf{QwQ-32B} & \textbf{Qwen3-32B} \\
\midrule[0.3mm]
Content Filling Requirement & 0.600 & 0.500 & 0.533 & 0.433 & \textbf{0.900} \\
\cline{1-6} 
Chief Complaint Filling Requirement & \textbf{1.000} & 0.971 & \textbf{1.000} & 0.914 & 0.814 \\
\cline{1-6} 
Personal History Filling Requirement & \textbf{0.643} & 0.143 & 0.571 & 0.429 & 0.143 \\
\cline{1-6} 
Family History Filling Requirement & \textbf{1.000} & 0.286 & 0.714 & 0.700 & 0.286 \\
\cline{1-6} 
Marital and Childbearing History Filling Requirement & \textbf{0.943} & 0.157 & 0.571 & 0.571 & 0.143 \\
\cline{1-6} 
Past Medical History Filling Requirement & \textbf{0.686} & 0.314 & 0.214 & 0.143 & 0.043 \\
\cline{1-6} 
Present Illness History Filling Requirement & 0.971 & 0.914 & \textbf{1.000} & \textbf{1.000} & \textbf{1.000} \\
\cline{1-6} 
Content Professionalism Requirement & \textbf{0.881} & 0.711 & 0.862 & 0.792 & 0.746 \\
\cline{1-6} 
Prohibit Outputting Extra Content & \textbf{0.773} & 0.713 & 0.677 & 0.420 & 0.547 \\
\cline{1-6} 
Prohibit Copying System Content & \textbf{0.917} & 0.793 & 0.757 & 0.577 & 0.747 \\
\cline{1-6} 
Dish Recommendation Quantity Requirement & 0.700 & \textbf{1.000} & 0.967 & \textbf{1.000} & 0.933 \\
\cline{1-6} 
Dish Recommendation Reason Explanation & \textbf{1.000} & \textbf{1.000} & \textbf{1.000} & \textbf{1.000} & 0.933 \\
\cline{1-6} 
Dish Recommendation Ingredients Requirement & 0.600 & \textbf{0.700} & 0.467 & \textbf{0.700} & 0.333 \\
\cline{1-6} 
User Health Analysis and Answer Requirement & \textbf{1.000} & \textbf{1.000} & \textbf{1.000} & \textbf{1.000} & 0.667 \\
\cline{1-6} 
Diet and Exercise Advice with Dish Recommendation & 0.833 & 0.967 & \textbf{1.000} & 0.867 & 0.933 \\
\bottomrule[0.5mm]
\end{tabular}
\caption{SSR Metrics on Industrial Applications Data in Thinking Mode.}
\label{tab:SSR Values for Each Model under Medical Scenario Constraints in Thinking Mode}
\end{table*}


\end{document}